\documentclass{article} % For LaTeX2e
\usepackage{iclr2027_conference,times}
\usepackage[T1]{fontenc}

\usepackage{amsmath,amsfonts,bm}

\def\eqref#1{equation~\ref{#1}}
\def\1{\bm{1}}

\DeclareMathAlphabet{\mathsfit}{\encodingdefault}{\sfdefault}{m}{sl}
\SetMathAlphabet{\mathsfit}{bold}{\encodingdefault}{\sfdefault}{bx}{n}

\usepackage[nohints]{minitoc}
\usepackage{url}
\usepackage{amsmath}
\usepackage{amssymb}
\usepackage{mathtools}
\usepackage{amsthm}
\usepackage{algorithm}
\usepackage{algpseudocode}
\usepackage{graphicx}
\usepackage{booktabs}
\usepackage{multirow}
\usepackage{soul}
\usepackage[table]{xcolor}
\usepackage{caption}
\usepackage{wrapfig}
\usepackage[letterpaper,textwidth=6.5in,textheight=9in,top=1in,headheight=32pt,headsep=25pt,footskip=30pt]{geometry}
\usepackage[breakable]{tcolorbox}
\usepackage{fontawesome5}

\definecolor{strattitleblue}{RGB}{25,70,135}
\definecolor{stratboxblue}{RGB}{230,242,255}
\definecolor{stratlinkblue}{RGB}{50,120,200}
\definecolor{straticonred}{RGB}{200,60,60}
\definecolor{straticonpurple}{RGB}{120,80,160}
\definecolor{straticonblue}{RGB}{60,100,180}
\newtcolorbox{strattitlebox}{
    colback=stratboxblue,
    colframe=stratboxblue,
    arc=4pt,
    boxrule=0pt,
    left=12pt,
    right=12pt,
    top=10pt,
    bottom=10pt,
    width=\textwidth,
    breakable,
}

\newenvironment{stratabstract}{%
    \vspace{0.2cm}%
    \noindent\textbf{Abstract:}\par\vspace{0.35em}%
    \setlength{\parskip}{0.4em}%
    \setlength{\parindent}{0pt}%
}{%
    \par\vspace{0.55cm}%
}

\newcommand{\stratkeywords}[1]{%
    \vspace{0.1cm}%
    \noindent\textbf{Keywords:}~#1\par
    \vspace{0.4cm}%
}

\newcommand{\stratgithubrepo}[1]{%
    \vspace{0.1cm}%
    \noindent{\small\textcolor{straticonpurple}{\faIcon{github}}~\textbf{Github Repo:}~{\hypersetup{urlcolor=magenta}\url{#1}}}\par
}
\newcommand{\stratcontact}[1]{%
    \vspace{0.1cm}%
    \noindent{\small\textcolor{straticonblue}{\faIcon{envelope}}~\textbf{Contact:}~~{\hypersetup{urlcolor=stratlinkblue}#1}}\par
}

\renewcommand{\headrulewidth}{0.4pt}
\renewcommand{\footrulewidth}{0pt}
\AtBeginDocument{\setlength{\headwidth}{\textwidth}}

\fancypagestyle{stratfirstpage}{%
    \fancyhf{}%
    \fancyhead[L]{%
        \includegraphics[height=14pt,keepaspectratio]{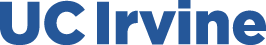}\hspace{14pt}%
        \includegraphics[height=14pt,keepaspectratio]{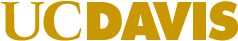}%
    }%
    \fancyhead[R]{\textsc{STRAT}}%
    \renewcommand{\headrulewidth}{0.4pt}%
    \renewcommand{\footrulewidth}{0pt}%
}

\definecolor{mydarkblue}{rgb}{0,0.08,0.45}
\usepackage[colorlinks=true,
    linkcolor=mydarkblue,
    citecolor=mydarkblue,
    filecolor=mydarkblue,
    urlcolor=mydarkblue]{hyperref}

\title{The Geometry of Logic: Stratification Induces\\[0.12cm]Semantic Structure and Robust Reasoning}

\author{\textbf{Cristina V. Lopes}$^{1,*}$\quad \textbf{Yuangang Li}$^{1,*,\ddagger}$\quad \textbf{Justin Tian Jin Chen}$^{2}$\\[0.1cm]
\textbf{Alberto Krone-Martins}$^{1}$\quad \textbf{Iris Ma}$^{1}$\quad \textbf{Md Rakib Hossain Misu}$^{1}$
}

\makeatletter
\renewcommand{\maketitle}{%
    \begin{center}
    {\fontsize{16}{20}\selectfont\bfseries\color{strattitleblue}%
    \raisebox{-0.28\height}{\includegraphics[height=26pt]{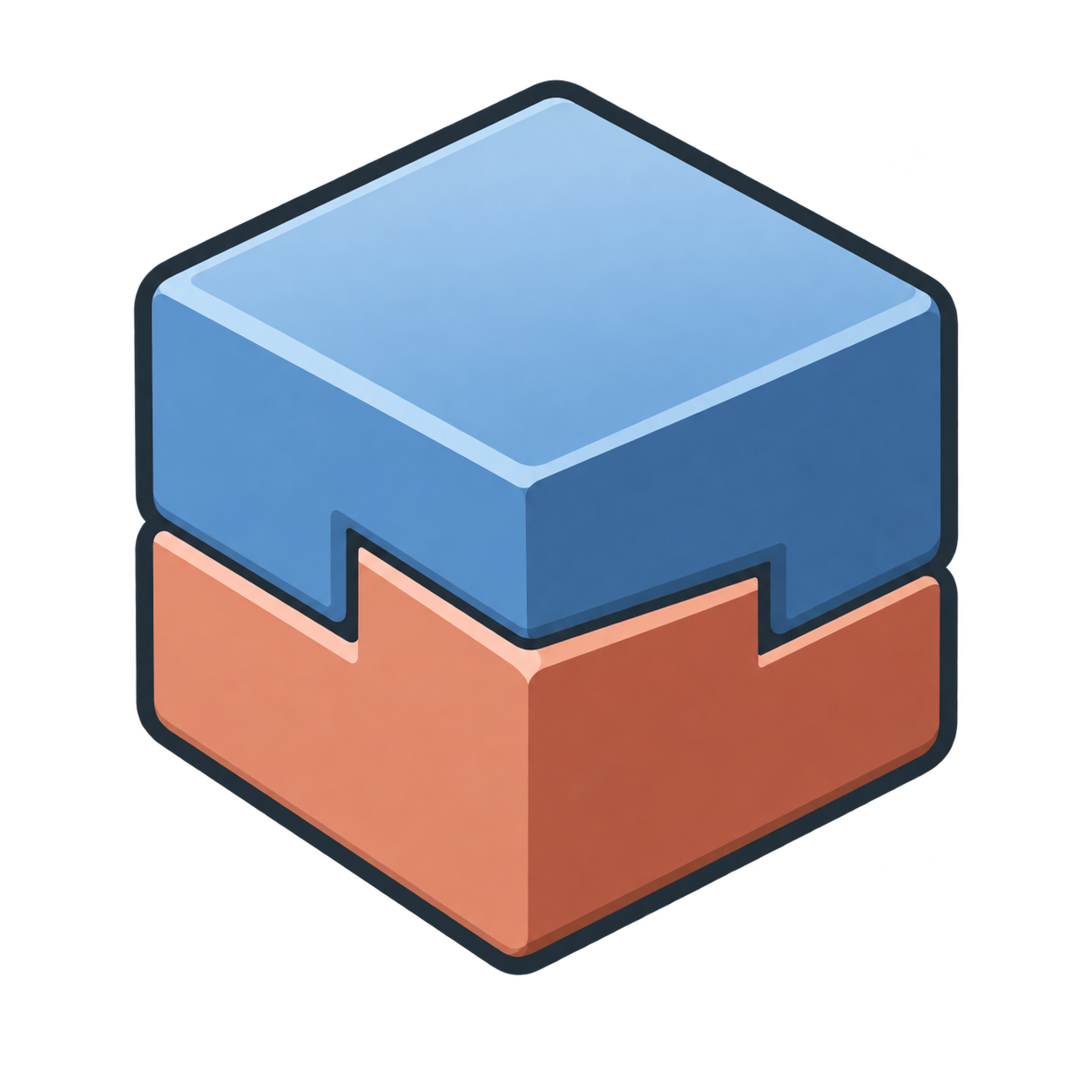}}\hspace{0.35em}\@title\par}
    \end{center}
    \vspace{0.3cm}
    \begin{center}
    \@author
    \end{center}
    \vspace{0.1cm}
    \begin{center}
    \footnotesize
    $^{1}$ University of California, Irvine\quad $^{2}$ University of California, Davis
    \end{center}
}
\makeatother

\iclrfinalcopy
\begin{document}

\doparttoc
\faketableofcontents
\thispagestyle{stratfirstpage}
\begin{strattitlebox}
\maketitle
\begin{stratabstract}
Transformer-based language models perform well on symbolic tasks, yet it remains unclear whether they learn generalizable rules or rely on statistical shortcuts. Mechanistic studies link algorithmic behavior to structured internal representations, motivating the hypothesis that robust reasoning benefits from separating values from the types that control their manipulation. Can making this separation an architectural primitive improve the learnability and generalization of logical mechanisms? We introduce \textbf{STRAT} (\textbf{ST}ratified \textbf{R}egisters \textbf{A}nd \textbf{T}ypes), which partitions the residual stream into orthogonal Data and Type subspaces and uses Type-based attention and gating to govern Data transformations. Controlled arithmetic ablations identify three failure modes associated with data-control interference: the Linear Trap, Gradient Wall, and Open Gate Trap. Mechanistic analysis reveals interpretable logical structure, and in arithmetic, STRAT reduces median OOD error 35-fold relative to a Transformer baseline. On each of 11 datasets spanning 10 tasks, STRAT outperforms the Transformer baseline in mean accuracy, by 26 percentage points on average, with both models trained from 10 base examples per dataset using identical task-specific augmentation where applicable. Under distribution shift, STRAT's mean accuracy drops by only 2.39 percentage points, compared with 11.75 for the Transformer.

\end{stratabstract}
\stratkeywords{Model Architecture, Representation Learning, Logical Reasoning, Mechanistic Interpretability}
% \stratpaperdate{\today}
\stratgithubrepo{https://github.com/crista/strat}
\stratcontact{\href{mailto:yuanganl@uci.edu}{yuanganl@uci.edu}}
\end{strattitlebox}
\begingroup
\renewcommand{\thefootnote}{\fnsymbol{footnote}}
\footnotetext[1]{These authors contributed equally. $^{\ddagger}$ Corresponding author.}
\endgroup

\vspace{-0.1in}
\begin{figure}[!h]
    \centering
    \includegraphics[width=1\textwidth]{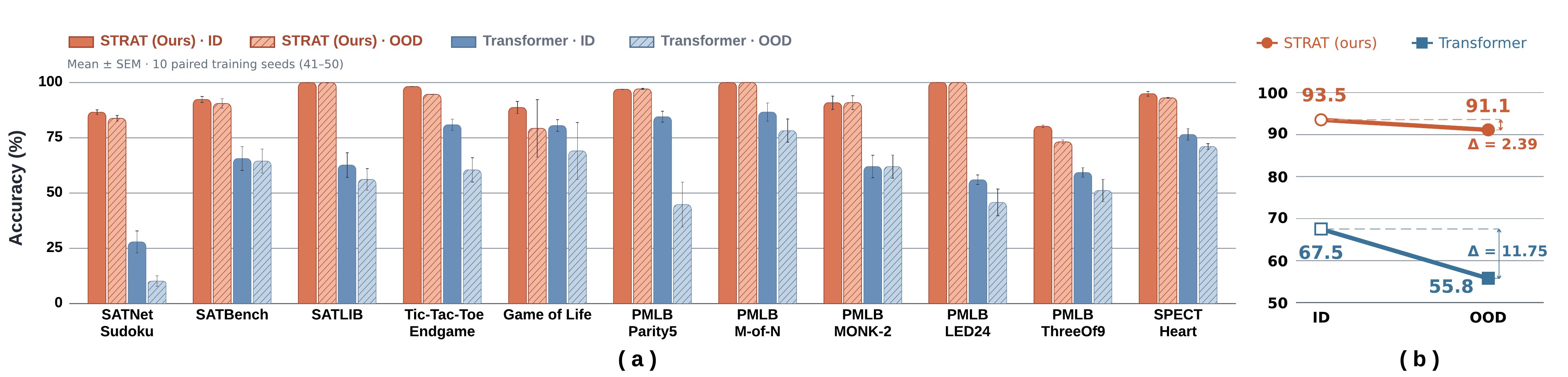}
    \vspace{-0.3in}
    \captionsetup{font=footnotesize} 
    \caption{Mean accuracy across 11 datasets spanning 10 tasks, with both STRAT and the Transformer baseline trained from only 10 base examples per dataset: \textbf{(a)} ID and OOD accuracy for each dataset; \textbf{(b)} macro-averaged accuracy under distribution shift. }
    \label{fig:main_results}
\end{figure}

\vspace{-0.2in}

\section{Introduction}
\vspace{-0.1in}
Transformer-based models have demonstrated an impressive fluency in symbolic domains, generating code and solving arithmetic problems with high apparent fidelity \citep{mcleish2024transformers,lee2024teaching,li2022alphacode,cho2025arithmetic}. Yet, a fundamental question regarding their internal mechanics remains unresolved \citep{nikankin2025arithmetic,mirzadeh2025gsmsymbolic}: Does the Transformer architecture truly reason, or does it merely approximate logic through massive statistical correlation?

Mechanistic interpretability research suggests that the answer may be ``both,'' depending on the stage of training. Recent findings indicate that Transformer models can transition from rote memorization to genuine algorithmic execution, a phenomenon known as \textit{grokking} \citep{power2022grokking,nanda2023progress}. This transition often coincides with the formation of discrete circuits, such as Induction Heads \citep{olsson2022incontext}, which implement copy-paste logic, and the emergence of internal world models that track logical state independent of surface statistics \citep{li2023othello}.

A unifying hypothesis for these phenomena is the \textit{Linear Representation Hypothesis} \citep{elhage2022toy, park2024the}, which posits that models represent distinct concepts as orthogonal directions in activation space. We extend this view to put forward the hypothesis that robust reasoning relies on a specific geometric configuration: the \textbf{stratification} of the embedding space into the basic building blocks of formal logic such as \textit{Values} (the variable content being manipulated) and \textit{Types} (the control signals governing the values). In standard Transformers, this separation must be learned against the gradient noise of high-variance data, potentially explaining the massive data requirements and abrupt phase transitions observed in training.

Taking a constructive approach, in this work, we ask: \textbf{What if this stratification were an architectural primitive rather than an emergent property?}

In order to explore this question, we step back from large transformer-based multi-purpose models, and focus on small ones whose purpose is to perform relatively simple but highly logical tasks like arithmetic and Sudoku. Our goal is to study how logic can emerge in neural models.

We introduce \textbf{STRAT} (\textbf{STrat}ified \textbf{R}egisters \textbf{A}nd \textbf{T}ypes), an architecture that explicitly partitions the residual stream into mutually exclusive subspaces for Data and Types. By structurally enforcing the ``geometric hygiene,'' we observe a radical shift in learning dynamics. Specifically, our contributions are as follows:
\vspace{-0.05in}
\begin{enumerate}
    \item \textbf{The Failure Taxonomy.} Through a rigorous ablation study, we demonstrate that Transformers fail to learn the tasks not due to insufficient scale or optimization noise, but due to \textbf{Data-Control Interference}. We identify three distinct failure modes: the \textit{Linear Trap}, the \textit{Gradient Wall}, and the \textit{Open Gate Trap}.
    
    \item \textbf{Stratified Attention.} We introduce a dual-stratum architecture that enforces orthogonality between Data (Values) and Control (Types), creating a protected ``Gradient Highway'' that lets the model learn discrete logical gates despite high-magnitude numerical payloads.

    \item \textbf{Topological Locking.} We demonstrate that STRAT converges to the correct logical mechanism from only $N=10$ training examples. While baselines default to linear interpolation (OOD Error $>7000$), STRAT locks into the correct geometric topology, achieving a $35\times$ reduction in median error and generalizing to out-of-distribution ranges.

    \item \textbf{Empirical Evaluation.} We evaluate STRAT on 11 datasets spanning 10 tasks, achieving a mean accuracy advantage of 26 percentage points over the Transformer baseline. Under distribution shift, STRAT maintains stable performance, with mean accuracy dropping by only 2.39 percentage points, compared with 11.75 for the Transformer (see Figure~\ref{fig:main_results})
\end{enumerate}

\vspace{-0.1in}
\section{The Model}
\label{sec:model}
\vspace{-0.1in}
In this work, we use a standard Transformer architecture optimized for algorithmic stability. We employ widely adopted modern components -- specifically relative attention and gated activations -- which, as we will show, provide the necessary primitives for symbolic manipulation.

\vspace{-0.1in}
\subsection{Representation: The Embedding Space}
\vspace{-0.1in}
Tokens $S = (t_1, \dots, t_T)$ are mapped to $\mathbb{R}^d$ via an embedding matrix $E$. We forgo absolute positional embeddings, relying strictly on relative biases in the attention layer to handle sequence order.
\vspace{-0.1in}
\subsection{Binding: The Attention Mechanism}
\vspace{-0.1in}
The Attention Layer solves the \textit{Binding Problem}: associating operands (e.g., numbers) with their governing operators. We employ single-head self-attention with a linear distance penalty (ALiBi) to enforce locality:
\begin{equation}
\label{eq:baseline_attention}
A = \text{softmax}\left(\frac{(H_0 W_Q)(H_0 W_K)^T}{\sqrt{d_k}} + M + B\right)
\end{equation}
where $B_{i,j} = -\lambda |i - j|$ decays attention with distance. This shielding function provides a soft window, ensuring that tokens preferentially fetch state from their nearest neighbors while suppressing distant distractors.
The output is computed as $H_{attn} = H_0 + A (H_0 W_V)$.
\vspace{-0.1in}
\subsection{Conditional Execution: The Gated FFN}
\vspace{-0.1in}
The Feed-Forward Network utilizes a Gated Linear Unit (GLU) with a \textbf{Sigmoid} activation. This choice is structural: unlike ReLU or GELU, the bounded $(0, 1)$ range of the sigmoid allows the network to implement discrete boolean logic (Open/Closed).
\begin{equation}
\label{eq:baseline_glu}
H_{out} = H_{attn} + (\sigma(H_{attn} W_{gate} + b_{gate}) \odot (H_{attn} W_{val} + b_{val}))
\end{equation}
Here, the \textit{Gate} projection acts as a differentiable switch conditioning the flow of the \textit{Value} payload.
\vspace{-0.1in}
\subsection{Readout: Global Accumulation}
\vspace{-0.1in}
For arithmetic tasks, the final result is often an aggregation of local updates. We therefore use global sum pooling followed by a linear decoding:
$y = \sum_{t=1}^T (H_{out}[t] W_{out})$.
This pooling mechanism is permutation-invariant with respect to the sequence positions, forcing the model to solve the task by modifying the token states themselves rather than relying on a ``read'' head at the last position.

\vspace{-0.1in}
\section{The Geometric ALU: Existence Proof}
\label{sec:implementation}
\vspace{-0.1in}

\begin{wraptable}{r}{0.52\textwidth}
\vspace{-1.35em} 
\centering
\captionsetup{font=footnotesize} 
\caption{Dimension Assignments in the Residual Stream}
\label{tab:dimension_map}
\vspace{-0.12in} 
{\scriptsize % \footnotesize  \scriptsize
\begin{tabular}{cll}
\toprule
\midrule
\textbf{Index} & \textbf{Content} & \textbf{Functional Role} \\
\midrule
D0 & \texttt{Value} & Numerical magnitude (Payload) \\
D1--D4 & \texttt{Flags} & Type Indicators (IsNum, IsOp, etc.) \\
D5 & \texttt{Accu} & Conditional Storage \\
\midrule
\bottomrule
\end{tabular}
} 
\vspace{-1em} 
\end{wraptable}
We first establish that the architecture is capable of expressing symbolic reasoning by hand-constructing weights that instantiate an Arithmetic Logic Unit (ALU) using a single Transformer block. 
The geometric algorithm unfolds in four steps:
1. \textbf{Representation:} Tokens are mapped to a 6-dimensional orthogonal basis (Table \ref{tab:dimension_map}), separating numerical payloads ($D_0$) from boolean flags ($D_1-D_4$).
2. \textbf{Binding:} Numbers query preceding tokens to ``acquire'' their governing operator via attention.
3. \textbf{Gating:} A logic gate activates only for subtracted numbers, acting as a differentiable switch.
4. \textbf{Execution:} Numbers that flow through the negative gate are ``copied'' to a scratchpad dimension ($D_5$). Effectively, numbers to be subtracted are ``Copy-and-Pasted'' from $D_0$ to $D_5$ -- that is, in a nutshell, our geometric algorithm. The full implementation details, including weight matrices, are provided in \textbf{Appendix \ref{app:gALU}}.

\vspace{-0.1in}
\section{The Unlearnable Geometry of Logic}
\label{sec:unlearnable}
\vspace{-0.1in}
While Section ~\ref{sec:implementation} proves the Geometric ALU is \textit{capable} of symbolic reasoning, is this solution \textit{learnable}? We conducted an exhaustive ablation study to see if standard gradient descent could discover this topology from scratch.

\vspace{-0.1in}
\subsection{Experimental Protocol}
\label{sec:exp_protocol}
\vspace{-0.1in}
We trained 12 configurations of the 6-dimensional architecture on the signed-sum task for 2,000 steps (128k samples total), in batches of 64, 50 runs (seeds) per config. Configuration details are provided in Appx.~\ref{app:configurations}.
\textbf{Metrics:} We report the \textbf{25\textsuperscript{th}-percentile}, \textbf{Median}, and \textbf{75\textsuperscript{th}-percentile} of the Mean Absolute Error (MAE) to capture the variance between ``trapped'' and overfitted runs.

\textbf{In-Distribution (ID):} $x \in [1, 100]$. Baselines: \textit{Input Mean Absolute Deviation (MAD)} ($\approx 25.0$, guessing $y=x$) vs. \textit{Target MAD} ($\approx 90.1$, guessing mean).
\textbf{Out-of-Distribution (OOD):} $x \in [2000, 5000]$. Distinguishes logic from curve-fitting.

\vspace{-0.1in}
\subsection{Results: A Taxonomy of Failure}
\vspace{-0.1in}
Despite the existence of a perfect solution (0.00 error), \textbf{none} of the 600 learned models converged to the geometric mechanism. Table \ref{tab:failures} reveals that every attempt fell into one of three topological traps.

\begin{table*}[!t]
\centering
\caption{The Failure Taxonomy (N=50 runs/config). We report 25\%-ile / Median / 75\%-ile errors. }
\label{tab:failures}
\vspace{-0.1in}
\resizebox{\textwidth}{!}{%
{\scriptsize
\begin{tabular}{l l l ccc cc l}
\toprule
\midrule
& & & \multicolumn{3}{c}{\textbf{ID MAE} (Train)} & \multicolumn{2}{c}{\textbf{OOD MAE} (Test)} & \\
\cmidrule(lr){4-6} \cmidrule(lr){7-8}
\textbf{Cfg} & \textbf{Strategy} & \textbf{Hypothesis} & \textbf{25\%} & \textbf{Med} & \textbf{75\%} & \textbf{Med} & \textbf{75\%} & \textbf{Diagnosis / Mechanism} \\
\midrule
\textbf{Ref} & Hand-Coded & \textit{Existence Proof} & \textbf{0.0} & \textbf{0.0} & 0.0 & \textbf{0} & 0 & \textbf{Geometric Logic} \\
\midrule
\textbf{A} & Simplified Xformer & Standard Training & 3.5 & 15.5 & 38.2 & 5,977 & 10,809 & \textbf{Linear Trap} \\
\textbf{A$_s$} & A Scaled & Scale Invariance & 97.1 & 98.7 & 100.1 & 6,202 & 6,265 & \textbf{Linear Trap} \\
\textbf{A$_{ln}$} & LayerNorm & Internal Stability & 9.6 & 16.9 & 25.9 & 8,378 & 9,856 & \textbf{Feature Collapse} \\
\textbf{A$_{curr}$} & Curriculum & Complexity Sched. & 212.2 & 600.3 & 1736.1 & 2306 & 2,682 & \textbf{Linear Trap} \\
\midrule
\textbf{B$_{fl}$} & Frozen Logic & Sharp Gates & 25.9 & 58.7 & 69.2 & 5,949 & 6,715 & \textbf{Gradient Wall} \\
\textbf{B$_{fa}$} & Frozen Att & Perfect Routing & 4.9 & 6.2 & 8.7 & 11,817 & 13,060 & \textbf{Magnitude Leak} \\
\textbf{B$_{ortho}$} & Regularized & Orthogonality & 2.3 & 9.8 & 35.2 & 9,750 & 12,088 & \textbf{Linear Trap} \\
\midrule
\textbf{C$_g$} & Learned Gain & Self-Amplification & 4.9 & 13.5 & 36.1 & 7,997 & 12,036 & \textbf{Gradient Wall} \\
\textbf{C$_{sb}$} & Signal Boost & Fixed Gain ($\gamma$) & 10.8 & 45.5 & 49.3 & 5,613 & 7,228 & \textbf{Gradient Wall} \\
\textbf{B$_c$} & Reg + Boost & Synergy & 12.7 & 46.5 & 48.2 & 5,611 & 9,677 & \textbf{Gradient Wall} \\
\textbf{C$_b$} & Balanced & Operating Point & 1.2 & 1.8 & 2.9 & 11,916 & 12,221 & \textbf{Operating Point Overfit} \\
\textbf{C$_{bi}$} & Bias Init & Gradient Flow & 3.1 & 12.7 & 37.4 & 8,471 & 11,815 & \textbf{Open gate trap} \\
\midrule
\bottomrule
\end{tabular}%
}}
\vspace{-0.1in}
\end{table*}

The \textit{Linear Trap} reflects value-based shortcut fitting; the \textit{Gradient Wall} arises when saturated gates block learning; and the \textit{Open Gate Trap} leaves gates acting as linear pass-throughs. Even low ID error does not imply OOD generalization, supporting the need to isolate control signals from numerical payloads (Appx.~\ref{app:failure_analysis}).

\vspace{-0.1in}
\subsection{Conclusion: Simplicity Bias is Structural}
\vspace{-0.1in}
This ablation study provides an answer to the structural critique. 
\textbf{Normalization (A$_{ln}$)} failed to stabilize the features (ID MAE $16.9$). 
\textbf{Scaling (A$_s$)} failed to balance the gradients. 
\textbf{Curriculum (A$_{curr}$)} failed to guide the optimization.

The robustness of the failure confirms that \textbf{Simplicity Bias} in this domain is not a transient optimization pathology, but a structural inevitability of Unityped Transformers. In the optimization landscape, the Value stream ($x$) offers a shortcut that is linearly correlated with the target ($|y| \approx |x|$). Deep networks preferentially learn features with simple, linear error surfaces (the Value shortcut) while starving complex, non-linear features (the Logic gate). Without brute-force scale, the logic gate cannot be \textit{trained} to ignore the data; it must be \textit{architecturally isolated} from it.

\vspace{-0.1in}
\section{The STRAT Architecture}
\vspace{-0.1in}
\label{sec:strats}

% \begin{figure}[!t]
%     \centering
%     \includegraphics[width=\textwidth]{figures/strat.pdf}
%     \caption{Overview of the STRAT-Lite architecture with an arithmetic example ($12 + 3 - 5$).}
%     \label{fig:architecture}
%     \vspace{-0.1in}
% \end{figure}

Our diagnosis of the baseline failures reveals a fatal pathology: \textbf{Data-Control Interference}. Standard Transformers conflate \textit{Data} (variable payloads) and \textit{Control} (routing signals) into a single, unityped vector space.
To resolve this, we propose the \textbf{STRAT} architecture (\textbf{STrat}ified \textbf{R}egisters \textbf{A}nd \textbf{T}ypes). We explicitly partition the embedding space $\mathbb{R}^{d_{model}}$ into orthogonal subspaces that never interact via linear mixing.

\vspace{-0.1in}
\subsection{Input Representation: Scalar Injection}
\label{sec:strat_input}
\vspace{-0.1in}

Standard Transformers learn dense embeddings for all tokens. STRAT employs a hybrid embedding scheme that enforces stratification at the input level. Let the embedding space be partitioned into indices $I_{val} \subset \{1..d\}$ and $I_{type} \subset \{1..d\}$.

\textbf{1. The Value Stratum ($V$):} For numerical tokens, the raw scalar magnitude $v \in \mathbb{R}$ is injected directly into a reserved dimension ($D_0$). This input embedding is \textbf{non-learnable} (fixed identity), ensuring the model perceives the exact quantity $v$. While the input is fixed, the \textit{transformations} applied to it ($W_{val}$) are fully learnable.

\textbf{2. The Type Stratum ($T$):} Discrete tokens (operators) are mapped to embeddings restricted to the Type dimensions ($D_1..D_4$). We explore two initialization strategies:
\textbf{Fixed Types (Oracle):} We inject hand-coded orthogonal vectors (e.g., $e_1$ for Plus, $e_2$ for Minus) to test the routing mechanism in isolation (Sections \ref{sec:strat_lite}--\ref{subsec:mechanism}).
\textbf{Latent Types (Learned):} We initialize with random noise to test if the model can autonomously discover the semantic structure (Section \ref{sec:latent_discovery}).
% \begin{itemize}
%     \item \textbf{Fixed Types (Oracle):} We inject hand-coded orthogonal vectors (e.g., $e_1$ for Plus, $e_2$ for Minus) to test the routing mechanism in isolation (Sections \ref{sec:strat_lite}--\ref{subsec:mechanism}).
%     \item \textbf{Latent Types (Learned):} We initialize with random noise to test if the model can autonomously discover the semantic structure (Section \ref{sec:latent_discovery}).
% \end{itemize}

\vspace{-0.1in}
\subsection{STRAT-Lite: Disentangled Attention and Routing}
\label{sec:strat_lite}
\vspace{-0.1in}

\begin{figure}[!t]
    \centering
    \includegraphics[width=\textwidth]{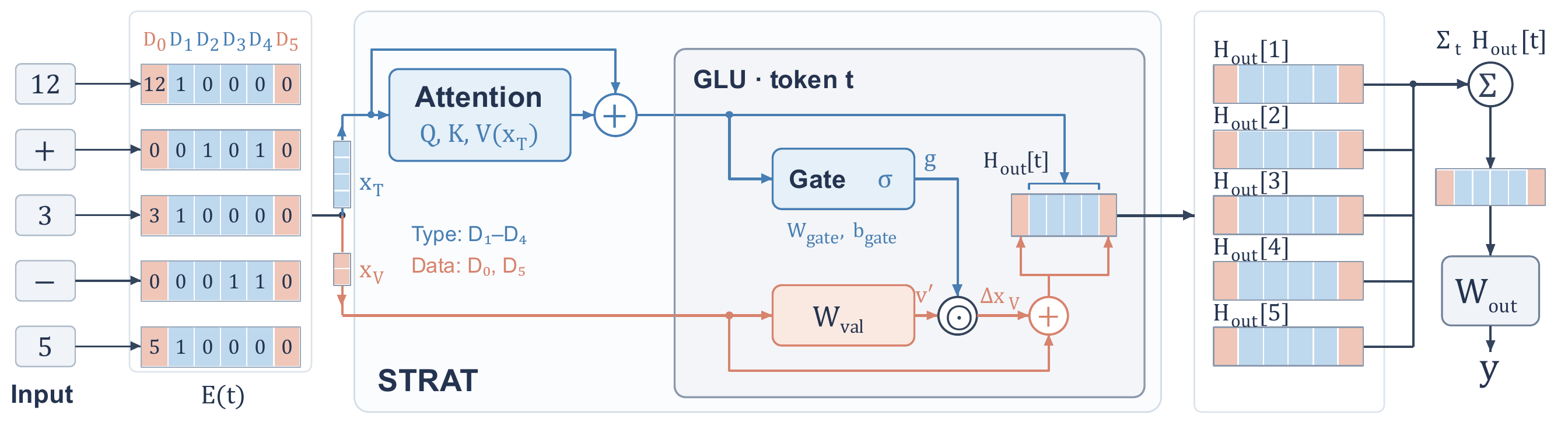}
    \caption{Overview of the STRAT-Lite architecture with an arithmetic example ($12 + 3 - 5$).}
    \label{fig:architecture}
    \vspace{-0.1in}
\end{figure}

To validate the core interaction between Data ($V$) and Types ($T$), we begin by analyzing a simplified instance, \textbf{STRAT-Lite}. 
% add by ken
Figure~\ref{fig:architecture} provides an overview of the architecture, using $12 + 3 - 5$ as an illustrative input.
The block update $x^{(l+1)} = \text{Block}(x^{(l)})$ modifies the standard Transformer block in two critical ways:

\vspace{-0.1in}
\subsubsection{Phase 1: Disentangled Control Attention}
\vspace{-0.1in}
We restrict the Attention mechanism to operate exclusively on the Type Stratum ($T$). Tokens update their internal state based on the \textit{types} of their neighbors without the numerical values ($V$) corrupting the routing logic.
\begin{equation}
\label{eq:strat_attention}
\begin{aligned}
    A &= \text{Softmax}\left(\frac{Q(x_{T}) K(x_{T})^\top}{\sqrt{d_{type}}} + \text{ALiBi}\right) \\
    \Delta x_{T} &= A \cdot V(x_{T}) \\
    x^{(l)}_{T} &\leftarrow x^{(l)}_{T} + \Delta x_{T} \quad \text{(Residual Update)}
\end{aligned}
\end{equation}
\textit{Note:} This update is applied \textit{only} to the Type indices. The Value indices remain unchanged during this phase ($x^{(l)}_{V} \leftarrow x^{(l)}_{V}$).

\vspace{-0.1in}
\subsubsection{Phase 2: The Topology-Aware GLU Router}
\vspace{-0.1in}

The Feed-Forward Network is replaced by a topology-aware Gated Linear Unit (GLU). The gating branch observes only the (now updated) Type Stratum to determine \textit{when} to act, while the value branch observes only the Value Stratum to determine \textit{what} to transmit.
\begin{equation}
\label{eq:strat_glu}
\begin{aligned}
    g &= \sigma(W_{gate} x_T + b_{gate}) & \text{(Control Logic)} \\
    v' &= W_{val} x_V & \text{(Data Transformation)} \\
    \Delta x_V &= g \odot v' & \text{(Gated Update)} \\
    x^{(l+1)}_{V} &\leftarrow x^{(l)}_{V} + \Delta x_{V}
\end{aligned}
\end{equation}
This additive update allows the model to perform \textit{destructive interference} (subtraction) on the original signal: $x_{new} = x_{old} + \Delta x$.

\textbf{Normalization Strategy:} Unlike Transformers, STRAT-Lite \textbf{omits Layer Normalization} to prevent the high-variance Value signal from suppressing the Type signal via variance normalization.

\vspace{-0.1in}
\subsection{Validation and Mechanism}
\label{subsec:mechanism}
\vspace{-0.1in}

We evaluated STRAT-Lite on the signed-sum task using fixed oracle types, and under the same experimental protocol used for the ablation study (Section~\ref{sec:exp_protocol}). STRAT-Lite achieves an excellent generalization (25\%-ile MAE 25.8; Median MAE 40.4 on OOD data), whereas standard Transformers fail catastrophically (MAE $>1000$, see Table~\ref{tab:failures}).

\textbf{Forensic Analysis: The Two Attractors}
To understand \textit{how} the model implements arithmetic, we inspected the learned weights of 50 successful seeds. We found that the architectural constraints force the model into one of two distinct topological ``attractors,'' both implementing a valid ``Cut-and-Paste'' logic.

\textbf{Attractor A (The Standard Topology):}
In some of seeds, the model discovers the direct subtraction mechanism:
\textbf{Gate:} The router learns a positive weight for the ``Minus'' type ($w_{gate} > 0$), opening the gate only when subtraction is required.
\textbf{Value:} The value weight is learned as negative ($w_{val} \approx -1$).
\textbf{Result:} The update $\Delta x = 1.0 \cdot (-x_{in})$ is added to the residual, effectively moving $-x_{in}$ to the accumulator.

% \begin{itemize}
%     \item \textbf{Gate:} The router learns a positive weight for the ``Minus'' type ($w_{gate} > 0$), opening the gate only when subtraction is required.
%     \item \textbf{Value:} The value weight is learned as negative ($w_{val} \approx -1$).
%     \item \textbf{Result:} The update $\Delta x = 1.0 \cdot (-x_{in})$ is added to the residual, effectively moving $-x_{in}$ to the accumulator.
% \end{itemize}

\textbf{Attractor B (The Mirror Topology):}
In the other seeds, the model discovers an algebraically equivalent but geometrically inverted solution:
\textbf{Gate:} The router learns a \textit{negative} weight for the ``Minus'' type ($w_{gate} < 0$), making the gate ``normally open'' but closed for Minus.
\textbf{Value:} The value weight is inverted ($w_{val} \approx 1$).
\textbf{Result:} The model effectively adds $x_{in}$ by default and learns to \textit{suppress} the addition when the type is Minus.

% \begin{itemize}
%     \item \textbf{Gate:} The router learns a \textit{negative} weight for the ``Minus'' type ($w_{gate} < 0$), making the gate ``normally open'' but closed for Minus.
%     \item \textbf{Value:} The value weight is inverted ($w_{val} \approx 1$).
%     \item \textbf{Result:} The model effectively adds $x_{in}$ by default and learns to \textit{suppress} the addition when the type is Minus.
% \end{itemize}

% The existence of these discrete, interpretable attractors confirms that STRAT does not merely fit a curve; it descends into a structured semantic minimum defined by the topology.
Appx.~\ref{app:alien_logic} provides representative learned weights and their analysis. The existence of these discrete, interpretable attractors confirms that STRAT does not merely fit a curve; it descends into a structured semantic minimum defined by the topology.

\vspace{-0.1in}
\subsection{Latent Type Discovery (Learned Types)}
\label{sec:latent_discovery}
\vspace{-0.1in}

Does this architecture require hand-engineered Type definitions? To test this, we ignored our fixed embeddings and initialized the Type Stratum ($T$) randomly ($\mathcal{N}(0, 0.02)$).

\vspace{-0.1in}
\subsubsection{Arithmetic}
\vspace{-0.1in}

We trained 10 models on the signed-sum task without explicit tags for Plus or Minus. They converged to the same MAE as the fixed-type models. As shown in Table \ref{tab:learned_types}, the model discovered the necessary logic by forcing the learned vectors for Addition and Subtraction to become antipodal ($v_+ \approx -v_-$), maximizing the decision margin for the GLU.

\begin{table}[htbp]
  \centering
  \begin{minipage}[t]{0.5\textwidth}
    \centering
    \captionsetup{font=footnotesize} 
    \caption{Learned Type Embeddings. Note the antipodal structure ($v_+ \approx -v_-$) between Plus and Minus.}
    \label{tab:learned_types}
    \vspace{-0.12in} 
    \resizebox{0.8\textwidth}{!}{% 
      \begin{tabular}{lrrrr}
      \toprule
      \midrule
      \textbf{Token} & \textbf{D1} & \textbf{D2} & \textbf{D3} & \textbf{D4} \\
      \midrule
      Literal & $0.55$ & $-0.25$ & $0.06$ & $0.06$ \\
      Plus & \textbf{$0.50$} & \textbf{$-0.59$} & \textbf{$-0.67$} & \textbf{$-0.51$} \\
      Minus & \textbf{$-0.80$} & \textbf{$0.81$} & \textbf{$0.83$} & \textbf{$0.79$} \\
      \midrule
      \bottomrule
      \end{tabular}%
    }
  \end{minipage}%
  \hfill
  \begin{minipage}[t]{0.48\textwidth}
    \centering
    \captionsetup{font=footnotesize} 
    \caption{Learned Centroids. Note the geometric collapse of the two Distractor categories.}
    \label{tab:centroids}
    \vspace{-0.11in} 
    \resizebox{\textwidth}{!}{%
      \begin{tabular}{lrrrr}
      \toprule
      \midrule
      \textbf{Category} & \textbf{D1} & \textbf{D2} & \textbf{D3} & \textbf{D4} \\
      \midrule
      Target (Cat 0) & $-0.15$ & $-0.18$ & $-0.19$ & $0.27$ \\
      Distractor A (Cat 1) & $0.27$ & $0.29$ & $0.31$ & $-0.44$ \\
      Distractor B (Cat 2) & $0.28$ & $0.29$ & $0.30$ & $-0.44$ \\
      \midrule
      \bottomrule
      \end{tabular}%
    }
  \end{minipage}
\vspace{-0.1in}
\end{table}

\vspace{-0.1in}
\subsubsection{Semantic Clustering (The ``Grocery List'')}
\vspace{-0.1in}
We further tested the model on a \textbf{Selective Summation} task with 15 distinct Item IDs drawn from three latent categories (Target, Distractor A, Distractor B).
\textbf{Results:} The model achieved 99.6\% OOD accuracy. Geometric analysis (Table \ref{tab:centroids}) reveals that the embeddings for the two Distractor categories collapsed into a shared manifold (Distance $< 0.01$), confirming that STRAT autonomously clustered tokens based on their functional role.

\vspace{-0.1in}
\subsubsection{PCA}
\vspace{-0.1in}

Figure~\ref{fig:latent_types} shows the visualization of the Principal Component Analysis (PCA) applied to the learned Type Stratum ($D_1-D_4$) after training. The geometry reveals two distinct mechanisms of semantic emergence:

\begin{wrapfigure}{R}{0.46\textwidth} 
  \vspace{-0.25in} 
  \centering
  \includegraphics[width=\linewidth]{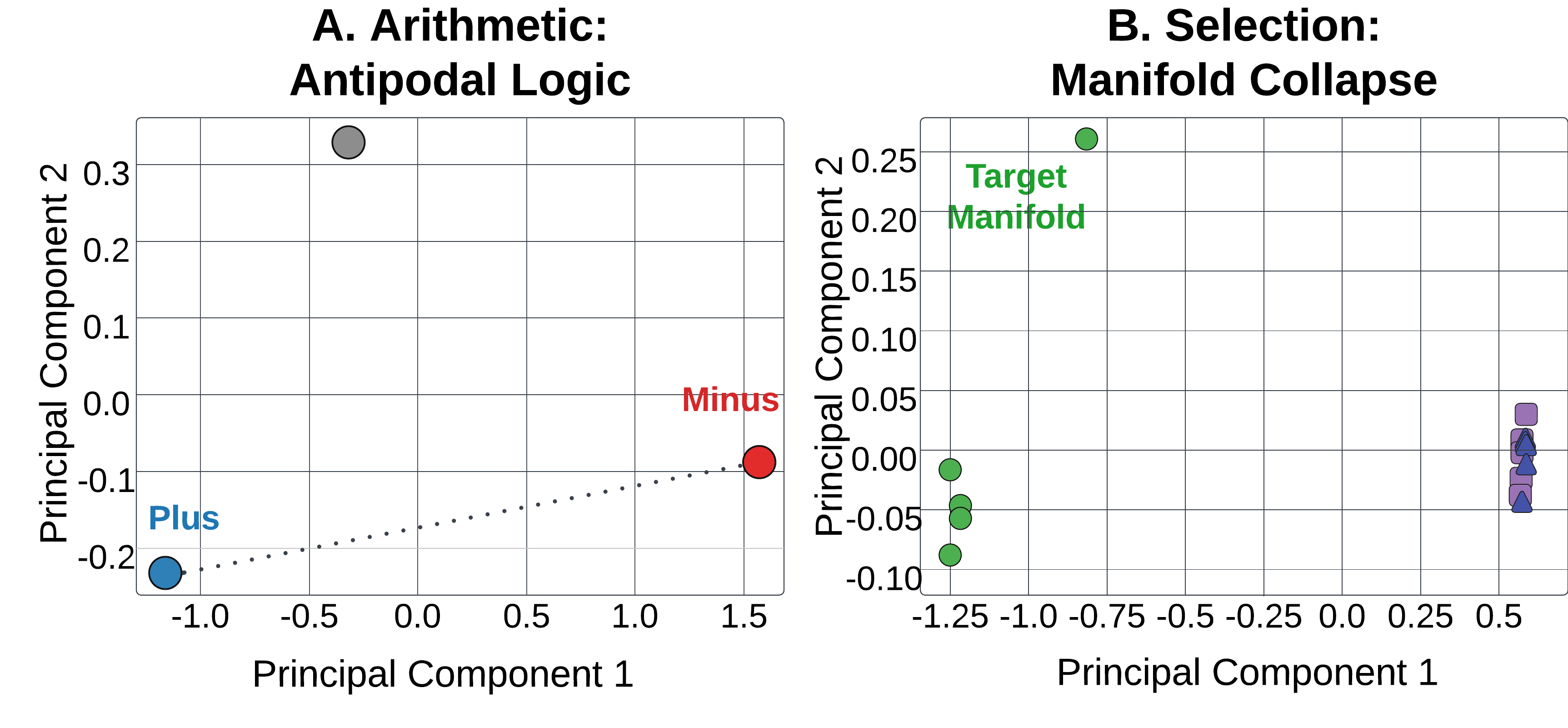}
  \vspace{-0.3in} \captionsetup{font=footnotesize} 
  \caption{Antipodal discovery. Left: arithmetic. Blue circle: plus; red circle: minus; grey circle: numbers. Right: selective sum. Green cuircles: target category; purple squares: distractor category A; blue triangles: distractor category B.}
  \label{fig:latent_types}
  \vspace{-1.5em} 
\end{wrapfigure}
\textbf{Panel A (Arithmetic):} The model spontaneously discovers an \textbf{antipodal arrangement} for the operator tokens. The Plus (Blue) and Minus (Red) vectors are positioned at opposite ends of the principal axis ($v_{+} \approx -v_{-}$), maximizing the orthogonality of their respective decision boundaries. The Literal token (Grey) remains in a neutral position near the origin, ensuring that raw numbers do not inherently trigger logic gates. This configuration confirms that the model implements signed arithmetic by learning diametrically opposed geometric directions for inverse operations.

\textbf{Panel B (Selection):} The model exhibits \textbf{manifold collapse}. Although Distractor A (Squares) and Distractor B (Triangles) represent distinct item IDs, the model maps them to an overlapping cluster in the latent space (Purple). In contrast, the Target items (Green) form a distinct, separated manifold. This proves that the model actively constructed a semantic Null Space, treating all non-target items as functionally identical regardless of their surface-level token identity.

\vspace{-0.1in}
\section{Topological Locking: Logic from N=10}
\label{sec:locking}
\vspace{-0.1in}

If a model possesses the correct structural prior, it should not require thousands of examples to approximate the solution. To test the limits of the architecture, we performed an extreme sample efficiency sweep.

\vspace{-0.1in}
\subsection{The Hyper-Overfitting Regime}
\vspace{-0.1in}
We instantiated \textbf{1,600 distinct learners} (400 random seeds per configuration). Each learner was provided with a theoretically minimal training dataset of size \textbf{N=10}: ten randomly sampled arithmetic sequences (e.g., ``$5 + 3 - 2$'') with values $x \in [0, 100]$.

We subjected the learners to a \textbf{Hyper-Overfitting Regime}: each model was trained on its tiny dataset for \textbf{2,000 epochs}, forcing it to memorize the training set perfectly (MSE $\approx 0$). To test robustness, we then evaluated on Out-of-Distribution sequences ($x \in [2000, 5000]$).

\vspace{-0.1in}
\subsection{Experimental Setup}
\vspace{-0.1in}
As baseline, we used Config A from the ablation study, and tested 4 variants of STRAT-Lite to determine the topology required for convergence: \textbf{\textit{(i)} STRAT-Lite (2D):} Compresses the Type Subspace to $\mathbb{R}^2$, forcing the model to find a planar separation between operators. \textbf{\textit{(ii)} STRAT-Lite (2D-Dual):} Adds a second Attention Head to the 2D bottleneck, testing whether parallel routing mechanisms facilitate convergence. \textbf{\textit{(iii)} STRAT-Lite (3D):} Expands the subspace to $\mathbb{R}^3$. Since there are three distinct token types (Literal, Plus, Minus), this dimension allows for an orthogonal simplex representation. \textbf{\textit{(iv)} STRAT-Lite (4D):} The high-capacity variant ($\mathbb{R}^4$) used in previous sections.

% \begin{itemize}
%     \item \textbf{\textit{(i)} STRAT-Lite (2D):} Compresses the Type Subspace to $\mathbb{R}^2$, forcing the model to find a planar separation between operators.
%     \item \textbf{\textit{(ii)} STRAT-Lite (2D-Dual):} Adds a second Attention Head to the 2D bottleneck, testing whether parallel routing mechanisms facilitate convergence.
%     \item \textbf{\textit{(iii)} STRAT-Lite (3D):} Expands the subspace to $\mathbb{R}^3$. Since there are three distinct token types (Literal, Plus, Minus), this dimension allows for an orthogonal simplex representation.
%     \item \textbf{\textit{(iv)} STRAT-Lite (4D):} The high-capacity variant ($\mathbb{R}^4$) used in previous sections.
% \end{itemize}

To strictly test the architectural inductive bias, we handicapped the STRAT models by forcing them to operate in the latent regime (requiring spontaneous type discovery). In contrast, we provided the Baseline (Config A) with explicit oracle supervision, injecting perfect one-hot type encodings directly into its input. 
To evaluate whether the model learned the subtraction rule versus merely fitting the training data, we define two accuracy thresholds based on the OOD MAE: \textbf{Logic Accuracy (MAE $<$ 100):} This measures the success of the routing mechanism. Given the magnitude of OOD inputs ($x \approx 3,500$), a failure to switch signs would result in massive errors ($\text{MAE} > 3,000$). An MAE below 100 proves the logic is correct, even if the gates are not perfectly saturated (the ``Sigmoid Tax''). \textbf{Strict Accuracy (MAE $<$ 20):} This measures the saturation of the Control Signal. It requires that the logic gates are driven to hard limits ($\sigma(x) \approx 1.0$), ensuring the signal passes with $>99\%$ fidelity.
% \begin{itemize}
%     \item \textbf{Logic Accuracy (MAE $<$ 100):} This measures the success of the routing mechanism. Given the magnitude of OOD inputs ($x \approx 3,500$), a failure to switch signs would result in massive errors ($\text{MAE} > 3,000$). An MAE below 100 proves the logic is correct, even if the gates are not perfectly saturated (the ``Sigmoid Tax'').
%     \item \textbf{Strict Accuracy (MAE $<$ 20):} This measures the saturation of the Control Signal. It requires that the logic gates are driven to hard limits ($\sigma(x) \approx 1.0$), ensuring the signal passes with $>99\%$ fidelity.
% \end{itemize}

\vspace{-0.1in}
\subsection{Result Analysis: The Linear Trap vs. Structure}
\vspace{-0.1in}
The results in Table \ref{tab:locking} reveal a fundamental dichotomy in learning dynamics:
\begin{table*}
\centering
% \caption{Topological Locking Results ($N=10$, 400 Seeds). \textbf{Strict Acc} ($<20$) denotes perfect circuit execution. \textbf{Logic Acc} ($<100$) denotes functional gating (allowing for Sigmoid saturation tax). The Baseline fails universally ($0\%$ acc). STRAT consistently achieves logic locking in $\approx 31\%$ of trials, with the top 10\% of runs achieving near-perfect precision (MAE $\approx 33$).}
\caption{Topological Locking Results ($N=10$, 400 Seeds). \textbf{Strict Acc} ($<20$) denotes perfect circuit execution. \textbf{Logic Acc} ($<100$) denotes functional gating with Sigmoid tax. The Baseline fails ($0\%$ acc). STRAT achieves $\approx 31\%$ logic locking, with the top 10\% achieving MAE $\approx 33$.}
\vspace{-0.1in}
\label{tab:locking}
\resizebox{0.8\textwidth}{!}{%
\begin{tabular}{l ccc cc}
\toprule
\midrule
& \multicolumn{3}{c}{\textbf{OOD MAE Distribution} (Lower is Better)} & \multicolumn{2}{c}{\textbf{Success Rates} (Higher is Better)} \\
\cmidrule(lr){2-4} \cmidrule(lr){5-6}
\textbf{Model} & \textbf{10\%-ile} & \textbf{25\%-ile} & \textbf{Median} & \textbf{Strict ($<$20)} & \textbf{Logic ($<$100)} \\
\midrule
Baseline (Config A) & $5,072$ & $6,268$ & $9,046$ & $0\%$ & $0\%$ \\
\midrule
STRAT Flat (2D/1H) & $33$ & $80$ & $293$ & $5\%$ & $31\%$ \\
STRAT Flat MH (2D/2H) & $38$ & $97$ & $314$ & $2\%$ & $27\%$ \\
STRAT Compact (3D/1H) & $34$ & $77$ & $266$ & $3\%$ & $31\%$ \\
STRAT Std (4D/1H) & $\mathbf{33}$ & $\mathbf{77}$ & $\mathbf{258}$ & $4\%$ & $\mathbf{31\%}$ \\
\midrule
\bottomrule
\end{tabular}
}
\vspace{-0.3in}
\vspace{0.1in}
\end{table*}
\textbf{The Baseline Failure (Linear Trap):}
The simplified baseline Transformer (Config A) minimizes training loss by learning a linear mapping $f(x) \approx Wx + b$ that passes through the 10 training points. The failure is absolute: even the top 10\% of Baseline runs yield an OOD error of \textbf{5,072}, and the Strict/Logic accuracy is exactly $0\%$. This confirms that without stratification, the probability of spontaneously discovering a boolean gate in the presence of high-magnitude data is effectively zero.
\textbf{The STRAT Success (Topological Locking):}
In contrast, STRAT consistently achieves a $\approx 31\%$ logic success rate across variants. The metrics reveal a tiered success structure: \textbf{\textit{(i)} Strict Locking ($\approx 4\%$):} A small subset of models achieve MAE $<20$, indicating they found the optimal ``Cut-and-Paste'' circuit with hard saturation. \textbf{\textit{(ii)} Logic Locking ($\approx 31\%$):} A significant portion achieve MAE $<100$. These models correctly implement the subtraction logic but pay a small ``Sigmoid Tax'' due to softer gate saturation. \textbf{\textit{(iii)} Robust Precision:} The \textbf{10\%-ile MAE of 33} confirms that the best runs are extremely accurate.
% \begin{itemize}
%     \item \textbf{Strict Locking ($\approx 4\%$):} A small subset of models achieve MAE $<20$, indicating they found the optimal ``Cut-and-Paste'' circuit with hard saturation.
%     \item \textbf{Logic Locking ($\approx 31\%$):} A significant portion achieve MAE $<100$. These models correctly implement the subtraction logic but pay a small ``Sigmoid Tax'' due to softer gate saturation.
%     \item \textbf{Robust Precision:} The \textbf{10\%-ile MAE of 33} confirms that the best runs are extremely accurate.
% \end{itemize}

The similarity between variants (Flat vs. Std) suggests that locking is robust to dimensionality: as long as at least one dimension is orthogonal to the value stream (the Type Stratum), the Gradient Highway exists. The higher Median MAE ($\approx 260$) reflects the $N=10$ constraint: when the curriculum lacks operator diversity, the model cannot learn to close the gate. However, when the curriculum is valid, STRAT locks; the Baseline never does.

We also compare STRAT with  Neural Accumulator (NAC) and Neural Arithmetic Logic Units (NALU)~\citep{Trask2018} to assess generalization against specialized arithmetic methods. Under the same training budget, STRAT Std achieves a median OOD MAE of 258, more than 22-fold lower than NAC (5,819) and NALU (5,770). Appx.~\ref{app:arithmetic_baselines} provides the full comparison and implementation details.

\vspace{-0.1in}
\subsection{Why STRAT Succeeds: The Gradient Highway}
\vspace{-0.1in}
STRAT enables this locking via two mechanisms: \textbf{\textit{(i)} Orthogonality:} The Value Stratum is blocked from the Attention mechanism. The router's gradients depend \textit{only} on the Type embeddings, creating a noise-free ``Gradient Highway'' for learning logic. \textbf{\textit{(ii)} Constraint:} The model cannot mix $D_0$ (Value) and $D_1$ (Type). To reduce loss, it is forced to use the only available degree of freedom: the GLU gate. This effectively deletes the linear interpolation solution from the hypothesis space.

% \begin{enumerate}
%     \item \textbf{\textit{(i)}Orthogonality:} The Value Stratum is blocked from the Attention mechanism. The router's gradients depend \textit{only} on the Type embeddings, creating a noise-free ``Gradient Highway'' for learning logic.
%     \item \textbf{\textit{(ii)}Constraint:} The model cannot mix $D_0$ (Value) and $D_1$ (Type). To reduce loss, it is forced to use the only available degree of freedom: the GLU gate. This effectively deletes the linear interpolation solution from the hypothesis space.
% \end{enumerate}
\vspace{-0.1in}
\section{STRAT Evaluation Across 10 Diverse Tasks}
\label{sec:diverse_tasks}
\vspace{-0.1in}
\definecolor{stratTableBody}{HTML}{FCEDE7}
\definecolor{stratTableHead}{HTML}{F5D1C4}
\definecolor{stratTableAccent}{HTML}{A84A31}
\definecolor{transformerTableBody}{HTML}{EFF4F8}
\definecolor{transformerTableHead}{HTML}{D6E2EE}
\definecolor{transformerTableAccent}{HTML}{537598}
\definecolor{differenceTableBody}{HTML}{EAF4F6}
\definecolor{differenceTableAccent}{HTML}{1B6784}
\begin{table*}[t]
\centering
\renewcommand{\arraystretch}{1.1}
\setlength{\tabcolsep}{7pt}
\setlength{\aboverulesep}{0pt}
\setlength{\belowrulesep}{0pt}
\newcommand{\taskcell}[2]{%
\begin{tabular}[c]{@{}c@{}}
  \textit{#1}\\[-2pt]
  #2
\end{tabular}%
}
\newcommand{\datasetcell}[2]{#1 \mbox{#2}}
\caption{
We report accuracy (\%) across 11 dataset evaluations as mean $\pm$ standard error of the mean (SEM) over 10 seeds. The difference columns use STRAT minus Transformer: ID accuracy ($\Delta$ ID Acc.),and OOD accuracy ($\Delta$ OOD Acc.). The OOD Acc. columns include the OOD performance and the difference relative to the ID accuracy in parentheses (\textcolor{green}{green} for increase, \textcolor{red}{red} for decrease).
}
\vspace{-0.1in}
\label{tab:task_overview}
\resizebox{\textwidth}{!}{%
\begin{tabular}{@{} cc@{\hspace{12pt}}
>{\columncolor{stratTableBody}[0pt][0pt]\hspace{4pt}}c<{\hspace{4pt}}@{}
>{\columncolor{stratTableBody}[0pt][0pt]\hspace{4pt}}c<{\hspace{4pt}}@{\hspace{6pt}}
>{\columncolor{transformerTableBody}[0pt][0pt]\hspace{4pt}}c<{\hspace{4pt}}@{}
>{\columncolor{transformerTableBody}[0pt][0pt]\hspace{4pt}}c<{\hspace{4pt}}@{\hspace{6pt}}
>{\columncolor{differenceTableBody}[0pt][0pt]\hspace{4pt}}c<{\hspace{4pt}}@{}
>{\columncolor{differenceTableBody}[0pt][0pt]\hspace{4pt}}c<{\hspace{4pt}}@{}}
\toprule
\midrule
\multirow{2}{*}{\textbf{Task}}
& \multirow{2}{*}{\textbf{Dataset}}
& \multicolumn{2}{@{}>{\columncolor{stratTableHead}[0pt][0pt]}c@{\hspace{6pt}}}{\textbf{\textcolor{stratTableAccent}{STRAT (Ours)}}}
& \multicolumn{2}{@{}>{\columncolor{transformerTableHead}[0pt][0pt]}c@{\hspace{6pt}}}{\textbf{\textcolor{transformerTableAccent}{Transformer}}}
& & \\
\cmidrule(l{0pt}r{6pt}){3-4}
\cmidrule(l{0pt}r{6pt}){5-6}
& & \textbf{ID Acc. (\%)}
& \textbf{OOD Acc. (\%)}
& \textbf{ID Acc. (\%)}
& \textbf{OOD Acc. (\%)}
& \multirow{-2}{*}{\textbf{\textcolor{differenceTableAccent}{$\Delta$ ID Acc.}}}
& \multirow{-2}{*}{\textbf{\textcolor{differenceTableAccent}{$\Delta$ OOD Acc.}}} \\
\midrule

\taskcell{Constraint satisfaction}{Sudoku solving}
& \datasetcell{SATNet Sudoku}{\citep{wang2019satnet}}
& $\mathbf{86.58 \pm 1.08}$
& $\mathbf{83.80 \pm 1.30}$ (\textcolor{red}{-2.78})
& $27.92 \pm 4.97$
& $10.18 \pm 2.36$ (\textcolor{red}{-17.74})
& $\mathbf{+58.66 \pm 5.22}$
& $\mathbf{+73.62 \pm 3.24}$ \\
\midrule

\taskcell{Formula satisfaction classification}{Given-assignment satisfaction}
& \datasetcell{SATBench}{\citep{wei2025satbench}}
& $\mathbf{92.33 \pm 1.36}$
& $\mathbf{90.50 \pm 2.15}$ (\textcolor{red}{-1.83})
& $65.64 \pm 5.38$
& $64.48 \pm 5.40$ (\textcolor{red}{-1.16})
& $\mathbf{+26.69 \pm 5.58}$
& $\mathbf{+26.02 \pm 5.54}$ \\
\midrule

\taskcell{Formula satisfaction classification}{Given-assignment satisfaction}
& \datasetcell{SATLIB}{\citep{hoos2000satlib}}
& $\mathbf{99.98 \pm 0.02}$
& $\mathbf{99.96 \pm 0.04}$ (\textcolor{red}{-0.02})
& $62.67 \pm 5.58$
& $56.18 \pm 4.91$ (\textcolor{red}{-6.49})
& $\mathbf{+37.31 \pm 5.58}$
& $\mathbf{+43.78 \pm 4.91}$ \\
\midrule

\taskcell{Board classification}{Winner detection}
& \datasetcell{Tic-Tac-Toe Endgame}{\citep{tic-tac-toe_endgame_101}}
& $\mathbf{98.19 \pm 0.00}$
& $\mathbf{94.64 \pm 0.00}$ (\textcolor{red}{-3.54})
& $80.91 \pm 2.53$
& $60.49 \pm 5.47$ (\textcolor{red}{-20.42})
& $\mathbf{+17.28 \pm 2.53}$
& $\mathbf{+34.15 \pm 5.47}$ \\
\midrule

\taskcell{State-transition prediction}{Next-state prediction}
& \datasetcell{Game of Life}{\citep{strandgaard2023gameoflife}}
& $\mathbf{88.71 \pm 2.72}$
& $\mathbf{79.26 \pm 12.99}$ (\textcolor{red}{-9.45})
& $80.54 \pm 2.64$
& $69.08 \pm 12.89$ (\textcolor{red}{-11.46})
& $\mathbf{+8.16 \pm 3.61}$
& $\mathbf{+10.18 \pm 17.94}$ \\
\midrule

\taskcell{Parity concept learning}{Five-bit parity}
& \datasetcell{PMLB Parity5}{\citep{olson2017pmlb}}
& $\mathbf{96.94 \pm 0.06}$
& $\mathbf{97.12 \pm 0.25}$ (\textcolor{green}{+0.19})
& $84.56 \pm 2.48$
& $44.81 \pm 10.13$ (\textcolor{red}{-39.75})
& $\mathbf{+12.38 \pm 2.49}$
& $\mathbf{+52.31 \pm 10.19}$ \\
\midrule

\taskcell{Threshold concept learning}{M-of-N rule}
& \datasetcell{PMLB M-of-N}{\citep{olson2017pmlb}}
& $\mathbf{100.00 \pm 0.00}$
& $\mathbf{100.00 \pm 0.00}$ (0.00)
& $86.65 \pm 4.14$
& $78.23 \pm 5.24$ (\textcolor{red}{-8.42})
& $\mathbf{+13.35 \pm 4.14}$
& $\mathbf{+21.77 \pm 5.24}$ \\
\midrule

\taskcell{Categorical rule induction}{Exact-two classification}
& \datasetcell{PMLB MONK-2}{\citep{olson2017pmlb}}
& $\mathbf{90.86 \pm 3.04}$
& $\mathbf{90.92 \pm 3.07}$ (\textcolor{green}{+0.06})
& $62.01 \pm 5.13$
& $61.99 \pm 5.22$ (\textcolor{red}{-0.03})
& $\mathbf{+28.84 \pm 7.17}$
& $\mathbf{+28.93 \pm 7.29}$ \\
\midrule

\taskcell{Density classification}{LED segment density}
& \datasetcell{PMLB LED24}{\citep{olson2017pmlb}}
& $\mathbf{100.00 \pm 0.00}$
& $\mathbf{100.00 \pm 0.00}$ (0.00)
& $56.01 \pm 2.14$
& $45.76 \pm 6.06$ (\textcolor{red}{-10.25})
& $\mathbf{+43.99 \pm 2.14}$
& $\mathbf{+54.24 \pm 6.06}$ \\
\midrule

\taskcell{Majority classification}{Nine-bit majority}
& \datasetcell{PMLB ThreeOf9}{\citep{olson2017pmlb}}
& $\mathbf{80.18 \pm 0.57}$
& $\mathbf{73.12 \pm 0.88}$ (\textcolor{red}{-7.05})
& $59.30 \pm 2.17$
& $51.18 \pm 5.03$ (\textcolor{red}{-8.12})
& $\mathbf{+20.88 \pm 2.14}$
& $\mathbf{+21.94 \pm 5.39}$ \\
\midrule

\taskcell{Density classification}{SPECT feature density}
& \datasetcell{SPECT Heart}{\citep{spect_heart_95}}
& $\mathbf{94.91 \pm 1.05}$
& $\mathbf{93.05 \pm 0.15}$ (\textcolor{red}{-1.86})
& $76.50 \pm 2.63$
& $71.04 \pm 1.35$ (\textcolor{red}{-5.46})
& $\mathbf{+18.40 \pm 2.37}$
& $\mathbf{+22.01 \pm 1.34}$ \\
\midrule
\multicolumn{2}{c@{\hspace{12pt}}}{\textbf{Average}}
& $\mathbf{93.51}$
& $\mathbf{91.13}$ \textbf{(\textcolor{red}{-2.39})}
& $\mathbf{67.52}$
& $\mathbf{55.77}$ \textbf{(\textcolor{red}{-11.75})}
& $\mathbf{+26.00}$
& $\mathbf{+35.36}$ \\
\midrule
\bottomrule
\end{tabular}%
}
\vspace{-0.15in}
\end{table*}

Using predefined task-specific Type encodings, we compare STRAT with the Transformer on 11 datasets covering 10 tasks (Table~\ref{tab:task_overview}). In every comparison, STRAT and the Transformer receive the same observable inputs and training data. Within each dataset, we match network depth, attention-head count, feed-forward hidden widths, optimizer, learning rate, training duration, loss function, and checkpoint-selection rule. 
Both train on 10 base examples per dataset and the same augmentations, if any, derived from them.
STRAT uses fewer trainable parameters than the Transformer baseline in all 11 dataset evaluations (Table~\ref{tab:parameter_counts}).
We report mean accuracy and SEM over 10 paired runs (seeds 41--50), using example-level accuracy for single-label tasks and exact-board accuracy for SATNet Sudoku and Game of Life. 
Appx.~\ref{app:dataset_configurations} details the task-specific augmentation procedures, the Type/Data input encodings supplied to both models, and their respective configurations.

STRAT achieves higher mean accuracy than the Transformer in all 11 evaluations, with a macro-average advantage of 26 percentage points (Table~\ref{tab:task_overview}). These results support our hypothesis that explicit Data--Type stratification provides an effective inductive bias for generalization from limited examples, with benefits extending beyond arithmetic to diverse task families.

\vspace{-0.1in}
\paragraph{Testing under Distribution Shift.}
To test whether the gains above reflect logic learning rather than dataset-specific shortcut fitting, we evaluate both models without retraining on identical OOD sets that preserve the task rules. For example, Sudoku puzzles have more unevenly distributed clues, while SATBench clause sequences are extended to length 100 by repeating existing clauses. Appx.~\ref{app:dataset_configurations} provides the complete construction procedures.
STRAT achieves higher mean accuracy in all 11 OOD evaluations. Its macro-average accuracy drops by only 2.39 percentage points, compared with 11.75 for the Transformer. This robustness to structural shifts supports the hypothesis that STRAT learns transferable task logic with reduced reliance on dataset-specific shortcuts.

Moreover, on Sudoku, Tic-Tac-Toe, Game of Life, M-of-N, and MONK-2, all of which retain their original prediction targets, STRAT achieves performance comparable to that of current state-of-the-art methods while using only 10 base training examples per dataset; see Appx.~\ref{app:sota_original_targets} for details.

\vspace{-0.15in}
\section{Discussion}
\label{sec:discussion}

\vspace{-0.1in}
\subsection{The Geometry of Efficiency (Interpreting N=10)}
\vspace{-0.1in}
Perhaps the most striking result of this study is the phenomenon of \textit{Topological Locking} (Section \ref{sec:locking}), where the model converges to the correct logical rule from as few as $N=10$ examples.
Standard deep learning theory suggests that models require massive data to ``grok'' abstract rules~\citep{power2022grokking}. STRAT avoids this by altering the geometry of the hypothesis space. By enforcing stratification, we effectively delete the shortcut solution (linear curve-fitting) from the search space. The optimizer is not learning the logic in the traditional sense of carving a manifold; rather, it is sliding into the only remaining minima allowed by the topology. This suggests that \textit{Sample Efficiency is a derivative of Topological Constraints}.

\vspace{-0.1in}
\subsection{Emergent Semantics via Null Spaces}
\vspace{-0.1in}
In Section \ref{sec:latent_discovery}, STRAT did not merely route data based on pre-existing types; it actively constructed a type system. By collapsing the ``Distractor'' items into a shared null-space manifold (distance $< 0.01$), the model discovered the concept of an ``Ignore'' class.
This reverses the standard neuro-symbolic paradigm. Rather than grafting symbolic modules onto neural networks, STRAT demonstrates that we can \textit{induce} semantic structure simply by restricting information flow. The Type Subspace acts as a pressure vessel: shielded from the Value stream, gradient descent is forced to use it solely for low-variance abstractions.

\vspace{-0.15in}
\subsection{Value-Invariance and Safety}
\vspace{-0.1in}
A common assumption is that performance requires opacity. STRAT challenges this. By design, it guarantees \textbf{Value-Invariance}: the logic gate is mathematically incapable of being ``tricked'' by the magnitude of the payload.
This has profound implications for AI Safety. Standard Transformer neurons are polysemantic; a large number (e.g., $5000$) might accidentally trigger a logic neuron simply because it pushes the activation sum over a threshold. STRAT reduces the surface area for such failures to zero. We envision architectures as hybrid systems: standard \textit{System 1} layers for perceptual processing feeding into \textit{System 2} STRAT blocks for rigorous reasoning~\citep{Bengio2019}.

\vspace{-0.15in}
\subsection{The ``Sigmoid Tax''}
\vspace{-0.1in}
While STRAT eliminates semantic interference, it remains an analog approximation of digital logic. The persistent error in successful models (MAE $\approx 25$) represents the \textbf{Sigmoid Tax}: the asymptotic cost of simulating discrete switching with continuous functions. Future work will explore techniques to anneal these activation functions toward hard step functions (Heaviside) during the final stages of training, potentially crystallizing the learned analog circuit into a discrete, zero-error algorithm.
\vspace{-0.15in}
\section{Related Work}
\vspace{-0.15in}

\subsection{Transformers on Algorithmic Tasks}
\vspace{-0.15in}
Despite their dominance in language, Transformers struggle with precise arithmetic and OOD extrapolation~\citep{Saxton2019, Lake2018}. Prior work attributes this to positional failures, proposing Relative or Randomized Positional Encodings (RPE) to force locality~\citep{Shaw2018, Ruoss2023, Nye2021}.
However, our baseline results (Config A) show that even with perfect relative encodings (ALiBi), standard Transformers fail. We identify \textbf{Data-Control Interference} as the distinct failure mode: a pathology where high-variance payloads drown out low-variance control signals. This differs from the ``grokking'' phenomenon~\citep{power2022grokking}, as we show it is an architectural, not merely observational, barrier.

\vspace{-0.15in}
\subsection{Mechanistic Interpretability and Negative Heads}
\vspace{-0.15in}
Our forensic analysis of the Cut-and-Paste logic (Section \ref{subsec:mechanism}) contributes to mechanistic interpretability~\citep{elhage2021mathematical, Olah2020}. The `Cut' mechanism, routing a negative copy of a value to cancel it out, bears a striking resemblance to the Copy-Suppression Heads discovered in GPT-2~\citep{McDougall2023}. In LLMs, these heads suppress the logits of previous tokens to correct for naive copying. STRAT suggests that this suppression logic is not just a statistical correction but a fundamental geometric primitive for algorithms over data in residual streams.

\vspace{-0.15in}
\subsection{Type Theory and Parametricity}
\vspace{-0.15in}
From the perspective of Programming Language Theory (PLT), the standard Transformer operates on a flat, \textit{unityped} vector space~\citep{pierce2002types}. Dense linear layers perform a \textit{destructive mixing} operation ($y = \mathbf{w}^T (\mathbf{x}_{val} + \mathbf{x}_{type})$), collapsing the distinction between the \textbf{Term Level} (Values) and the \textbf{Type Level}.
STRAT resolves this by enforcing \textbf{Relational Parametricity}~\citep{Reynolds1983, Wadler1989}. By isolating the \textit{Value Stratum} as an opaque container, the logic becomes parametric with respect to the values: it can route the data, but it is architecturally forbidden from inspecting the \textit{bits} of the data itself. This mirrors the design of NALU~\citep{Trask2018, Madsen2020} but achieves the effect using  Transformer components rather than specialized ones.

\vspace{-0.15in}
\section{Conclusion}
\label{sec:conclusion}
\vspace{-0.15in}
We hypothesize that the reasoning capabilities attributed to scale are the result of implicit \textbf{Geometric Stratification}. By explicitly engineering this structure in STRAT, we demonstrated that core reasoning mechanisms can be achieved with extreme efficiency, rather than brute force. Our experiments revealed:
% \textbf{1. Geometry is Algorithm:} The model implemented task logic not by memorization, but by discovering specific topologies (e.g., antipodal arrangements).
% \textbf{2. Constraints Enable Generalization:} By restricting the search space to orthogonal strata, the model generalized from just $N=10$ examples.
% \textbf{3. Orthogonality Enables Logic:} A single set of weights successfully multiplexed contradictory programs, provided their execution contexts were geometrically orthogonal.
\vspace{-0.08in}
\begin{enumerate}
    \item \textbf{Geometry is Algorithm:} The model implemented task logic not by memorization, but by discovering specific topologies (e.g., antipodal arrangements).
    \item \textbf{Constraints Enable Generalization:} By restricting the search space to orthogonal strata, the model generalized from just $N=10$ examples.
    \item \textbf{Orthogonality Enables Logic:} A single set of weights successfully multiplexed contradictory programs, provided their execution contexts were geometrically orthogonal.
\end{enumerate}
\vspace{-0.08in}
We conclude that logic is not a symbolic rule set superimposed on a network, but the natural consequence of a stratified vector space. Future work should investigate whether the ``ghosts'' of these strata can be detected within the latent spaces of frontier LLMs.

\vspace{-0.1in}
\subsection*{AI use statement}
\vspace{-0.1in}
We used AI assistants to check grammar, polish our writing, draft sections of the paper, retrieve relevant literature, and assist with routine coding tasks. We reviewed all AI-assisted work, revised the text as needed, verified the cited sources for accuracy and relevance, and checked the code for correctness. We take full responsibility for the final content of this work.

\vspace{-0.1in}
\subsection*{Ethics statement}
\vspace{-0.1in}
This work studies architectural inductive biases for rule learning using synthetic tasks and publicly available benchmark datasets. We do not recruit human participants, collect new personal data, or use personally identifiable information. We do not anticipate significant ethical concerns arising from these experiments.

% \vspace{-0.1in}
% \subsection*{Reproducibility statement}
% \vspace{-0.1in}
% An implementation of the STRAT architecture is available in the anonymous repository at \url{https://anonymous.4open.science/r/strat-EB54/}.

\vspace{-0.1in}
\subsubsection*{Author Contributions}
\vspace{-0.1in}
Justin Tian Jin Chen, Alberto Krone-Martins, Iris Ma, and Md Rakib Hossain Misu are listed in alphabetical order by surname.

% \subsubsection*{Acknowledgments}
% Use unnumbered third level headings for the acknowledgments. All
% acknowledgments, including those to funding agencies, go at the end of the paper.

\bibliography{iclr2027_conference}
\bibliographystyle{iclr2027_conference}

\appendix
% \section{Appendix}
% You may include other additional sections here.
\clearpage
\appendix
\newpage
\renewcommand\thepart{}
\renewcommand\partname{}
\setcounter{secnumdepth}{-1}
\part{Appendix}
\setcounter{secnumdepth}{3}
\vspace{-3ex}
\setcounter{tocdepth}{2}
\parttoc
\newpage

%%%%%%%%%%%%%%%%%%%%%%%%%%%%%%%%%%%%%%%%%%%%%%%%%%%%%%%%%%%%%
%% Appendix A
%%%%%%%%%%%%%%%%%%%%%%%%%%%%%%%%%%%%%%%%%%%%%%%%%%%%%%%%%%%%%
\section{The Geometric ALU ``Program''}
\label{app:gALU}

This appendix provides the exact matrix configurations for the existence proof discussed in Section 
\ref{sec:implementation}.

\subsection{Dimension Map and Embeddings}

We utilize a sparse orthogonal basis in $\mathbb{R}^6$ (Table \ref{tab:dimension_map_app2}). $D_0$ serves as the continuous payload, while $D_1-D_4$ act as discrete property flags.

\begin{table}[h]
\centering
\caption{Dimension Assignments in the Residual Stream. }
\label{tab:dimension_map_app2}
\begin{tabular}{cll}
\toprule
\midrule
\textbf{Index} & \textbf{Identifier} & \textbf{Functional Role} \\
\midrule
D0 & \texttt{Value} & Numerical magnitude $n$ (Payload) \\
D1 & \texttt{Is\_Num} & Binary flag ($1.0$) for numerical literals \\
D2 & \texttt{Is\_Add} & Binary flag ($1.0$) for the ‘+’ operator \\
D3 & \texttt{Is\_Sub} & Binary flag ($1.0$) for the ‘-’ operator \\
D4 & \texttt{Is\_Op} & Generic flag ($1.0$) for any operator  \\
D5 & \texttt{Accu} & Reserved register for conditional storage \\
\midrule
\bottomrule
\end{tabular}
\end{table}

\begin{table}[h]
\centering
\caption{Sparse Token Embeddings $E(t)$}
\label{tab:embeddings}
\begin{tabular}{l|cccccc}
\toprule
\midrule
\textbf{Token} & \textbf{D0} & \textbf{D1} & \textbf{D2} & \textbf{D3} & \textbf{D4} & \textbf{D5} \\
\midrule
Literal ($n$)  & $n$ & $1$ & $0$ & $0$ & $0$ & $0$ \\
Plus ($+$)     & $0$ & $0$ & $1$ & $0$ & $1$ & $0$ \\
Minus ($-$)    & $0$ & $0$ & $0$ & $1$ & $1$ & $0$ \\
\midrule
\bottomrule
\end{tabular}
\end{table}

\subsection{Attention Weights}

The Attention Layer solves the \textit{Binding Problem} via an asymmetric ``Look-Back'' mechanism. We configure the Query/Key projections to establish a handshake between \texttt{Number} tokens and \texttt{Operator} tokens. The linear distance penalty (ALiBi) resolves collisions, ensuring that operands strictly attend to their nearest neighbor.

\textbf{Addressing \& Locality:} We configure the Query projection to map the Number Flag ($D_1$) to a handshake channel ($D_4$), and the Key maps the Operator Flag ($D_4$) to the same. A high gain ($G=10$) ensures numbers attend strongly to operators. Ties between multiple operators (e.g., $12 + 3 - 5$) are resolved by the linear distance penalty (ALiBi), $S_{ij} = (q_i \cdot k_j^T) - |i - j|$, which enforces a monotonic preference for the nearest neighbor.

\textbf{Fetching:} The Value projection copies the operator's flags ($D_2, D_3$) into the number's residual stream. The resulting state is a noisy superposition: a number token $t$ contains a strong signal from its nearest operator and weak ``leakage'' from distant ones.

$$W_Q = \begin{bmatrix}
0 & 0 & 0 & 0 & 0 & 0 \\
0 & 0 & 0 & 0 & \mathbf{10.0} & 0 \\ % D1 -> D4
0 & 0 & 0 & 0 & 0 & 0 \\
0 & 0 & 0 & 0 & 0 & 0 \\
0 & 0 & 0 & 0 & 0 & 0 \\
0 & 0 & 0 & 0 & 0 & 0 
\end{bmatrix}, \quad
W_K = \begin{bmatrix}
0 & 0 & 0 & 0 & 0 & 0 \\
0 & 0 & 0 & 0 & 0 & 0 \\
0 & 0 & 0 & 0 & 0 & 0 \\
0 & 0 & 0 & 0 & 0 & 0 \\
0 & 0 & 0 & 0 & \mathbf{10.0} & 0 \\ % D4 -> D4
0 & 0 & 0 & 0 & 0 & 0 
\end{bmatrix}$$

The Value projection $W_V$ copies the operator flags into the number's stream:
$$W_V = \begin{bmatrix}
0 & 0 & 0 & 0 & 0 & 0 \\
0 & 0 & 0 & 0 & 0 & 0 \\
0 & 0 & \mathbf{1.0} & 0 & 0 & 0 \\
0 & 0 & 0 & \mathbf{1.0} & 0 & 0 \\
0 & 0 & 0 & 0 & 0 & 0 \\
0 & 0 & 0 & 0 & 0 & 0 
\end{bmatrix}$$

\subsection{FFN Weights}

The attention output -- a superposition of the number's identity and the operator's type -- is then passed to the Gated Linear Unit (GLU). The GLU acts as a discriminator to filter attention noise. The gate is configured as a boolean AND circuit: it opens only if the token is a Number ($D_1$) \textit{and} attends to a Minus sign ($D_3$). Importantly, the gate weights for the Value dimension ($D_0$) are set to zero, rendering the logic \textbf{value-invariant}.

\textbf{The Gate ($W_{gate}$):} The gate targets the condition $\texttt{Is\_Num} \land \texttt{Is\_Sub}$. We set weights of $100.0$ for both $D_1$ (Num) and $D_3$ (Sub), with a bias of $-150.0$.
$$ \text{Gate} = \sigma(100 \cdot \texttt{Num} + 100 \cdot p_{sub} - 150) $$
For a number token ($\texttt{Num}=1$), this simplifies to $\sigma(100 \cdot p_{sub} - 50)$. If the nearest neighbor is Subtraction ($p_{sub} \approx 1.0$), the activation is $\sigma(50) \approx 1$. If it is Addition ($p_{sub} \approx 0$), the activation is $\sigma(-50) \approx 0$. Note that $W_{gate}$ has zero weights for $D_0$, rendering logic independent of magnitude.

\textbf{The Value ($W_{val}$):} The value matrix simply maps the numerical payload ($D_0$) to the accumulator dimension ($D_5$).

$$W_{gate} = 
\begin{bmatrix}
0 & 0 & 0 & 0 & 0 & 0 \\
0 & 0 & 0 & 0 & 0 & \mathbf{100.0} \\ % D1 -> D5
0 & 0 & 0 & 0 & 0 & 0 \\
0 & 0 & 0 & 0 & 0 & \mathbf{100.0} \\ % D3 -> D5
0 & 0 & 0 & 0 & 0 & 0 \\
0 & 0 & 0 & 0 & 0 & 0 
\end{bmatrix}, \quad
b_{gate} = [0, 0, 0, 0, 0, \mathbf{-150}]$$

The Value matrix $W_{val}$ copies the magnitude D0 to D5.
$$W_{val} = 
\begin{bmatrix}
0 & 0 & 0 & 0 & 0 & \mathbf{1.0} \\
0 & 0 & 0 & 0 & 0 & 0 \\
0 & 0 & 0 & 0 & 0 & 0 \\
0 & 0 & 0 & 0 & 0 & 0 \\
0 & 0 & 0 & 0 & 0 & 0 \\
0 & 0 & 0 & 0 & 0 & 0 
\end{bmatrix}$$

\subsection{Readout}

The output is computed via global sum pooling and projection $W_{out}$. $D0$ has the absolute values of all the numbers; $D5$ has the absolute values of the negative numbers, so we set weights $w_{D0}=1$ and $w_{D5}=-2$.
$$ y = \sum (1 \cdot D_0) + \sum (-2 \cdot D_5) $$

% \newpage

%%%%%%%%%%%%%%%%%%%%%%%%%%%%%%%%%%%%%%%%%%%%%%%%%%%%%%%%%%%%%
%% Appendix B
%%%%%%%%%%%%%%%%%%%%%%%%%%%%%%%%%%%%%%%%%%%%%%%%%%%%%%%%%%%%%
\section{Ablation Study Configurations}
\label{app:configurations}

To investigate whether the failure of dense networks to discover the geometric topology was due to specific hyperparameter choices or initialization strategies, we designed 12 distinct experimental configurations. These span the gamut of standard deep learning ``tricks,'' including signal boosting, logical pre-initialization, auxiliary loss penalties, and data normalization.

Table \ref{tab:config_details} summarizes the specific modifications applied to the base model for each configuration.
\begin{table*}[h]
\centering
\caption{Detailed hyperparameters for the 12 Ablation Configurations. All models share the same 6-dim architecture.}
\label{tab:config_details}
\resizebox{0.95\textwidth}{!}{%
\begin{tabular}{l l p{6cm} p{5cm}}
\toprule
\midrule
\textbf{Config} & \textbf{Name} & \textbf{Mechanism / Modification} & \textbf{Hypothesis Tested} \\
\midrule
\textbf{A} & Baseline & Standard Xavier initialization; Raw inputs. & Can standard GD solve it? \\
\textbf{A$s$} & Scaled & Inputs/Targets scaled (max-normalized). & Is magnitude the problem? \\
\textbf{A$_{ln}$} & LayerNorm & Internal LayerNorm & Feature Collapse (OOD Fail) \\
\textbf{A$_{curr}$} & Curriculum & Range $[0, 100] \to [0, 5000]$ & Linear Trap (Locked Early) \\
\midrule
\textbf{B$_{fl}$} & Frozen Logic & Logic weights fixed to ideal; others learned. & Can gradients flow through logic? \\
\textbf{B$_{fa}$} & Frozen Attn & Attention weights fixed to Oracle (Perfect). & Is the failure in Routing or Logic? \\
\textbf{B$_{ortho}$} & Value orthogonality & Loss penalty: $\lambda || w \cdot x ||$. & Can we punish value-dependence? \\
\midrule
\textbf{C$_{sb}$} & Signal Boost & Fixed pre-activation gain $\gamma=100$. & Is the gradient signal too weak? \\
\textbf{C$_g$} & Learned Gain & Learnable scalar gain $\alpha$ (init 1.0). & Can the model learn to boost signals? \\
\textbf{C$_b$} & Balanced & Signals balanced manually. & Does a valid operating point exist? \\
\textbf{C$_c$} & Combined & Config B$_{ortho}$ + Config C$_b$ & Do constraints + boosting help? \\
\textbf{C$_{bi}$} & Bias Init & Gate Bias = +1.0 & Dead gates? \\
\midrule
\bottomrule
\end{tabular}
}
\end{table*}

\subsection{Detailed Configuration Descriptions}

\textbf{Group A: Baselines and Normalization}
\begin{itemize}
    \item \textbf{Config A (Simple Baseline):} A simplified 6-dimensional Transformer block without input normalization or layer normalization. We utilize a single-head attention mechanism with \textbf{unscaled dot-product scoring} and a fixed linear decay bias ($B_{i,j} = -|i-j|$) to strictly enforce locality. The FFN utilizes a \textbf{Sigmoid-GLU activation} with learnable biases, while attention and readout projections are bias-free. Output is computed via \textbf{global sum-pooling} followed by a linear projection. \textit{Hypothesis Tested:} Can standard gradient descent, operating in a raw, unityped vector space, overcome the ``Simplicity Bias'' of linear correlations? This configuration serves as the control to determine if the failure to reason is intrinsic to the lack of stratification rather than specific optimization pathologies like vanishing gradients or covariate shift. \textit{Structural Note:} We employ \textbf{global sum-pooling} ($y = W_{out} \sum x_t$) rather than the standard last-token readout. This architectural choice is critical: it aligns the readout mechanism with the accumulation nature of the arithmetic task, forcing the model to modify token states directly rather than relying on a privileged ``read head'' at the end of the sequence. Additionally, the attention scores are \textit{unscaled} (omitting the $1/\sqrt{d_k}$ factor) given the low dimensionality ($d=6$).

    \item \textbf{Config A$_s$ (Scaled Inputs):} Identical to Config A, but with \textbf{Max-Normalization} applied to the numerical inputs and targets. We apply a fixed scalar gain of $\gamma = 1/5000 = 0.0002$ to the Value payload ($D_0$) and the regression targets, mapping the operating range from $[0, 5000]$ to $[0, 1]$. \textit{Hypothesis Tested:} Is the ``Linear Trap'' caused by the sheer magnitude imbalance between the Value channel ($x \approx 5000$) and the Control flags ($x \in \{0,1\}$)? By normalizing the payload to the same unit scale as the flags, we test if the failure to learn logic is simply a result of the optimizer struggling with heterogeneous scales (i.e., vanishing gradients on the control signals due to large value-driven updates). \textit{Structural Note:} Unlike standard normalization (e.g., BatchNorm) which relies on batch statistics, we employ a \textbf{fixed domain-scaling factor}. This ensures that the transformation is perfectly reversible and prevents the leakage of batch statistics into the inference logic. This places the Value neuron $D_0$ on the exact same magnitude footing as the Type neurons $D_1-D_5$ without altering the distribution's shape.
    
    \item \textbf{Config A$_{ln}$ (LayerNorm):} Similar to Config A, but we apply standard Layer Normalization in a \textbf{Pre-LN} configuration (before Attention and FFN blocks) to the $d=6$ residual stream. \textit{Hypothesis Tested:} This configuration tests whether the optimization instability (e.g., Gradient Walls) observed in baselines stems from internal covariate shift. \textit{Structural Note:} We acknowledge that applying LayerNorm to a low-dimensional space ($d=6$) containing a single high-variance ``Value'' neuron ($D_0$) alongside five low-variance ``Type'' neurons ($D_1-D_5$) represents a statistical extreme. This design is intentional; it serves as a rigorous test of \textbf{Unityped Normalization}. By forcing the numerical payload to share normalization statistics ($\mu, \sigma$) with the control signals, we explicitly test the hypothesis that coupled normalization destroys the absolute magnitude information required for arithmetic.

    \item \textbf{Config A$_{curr}$ (Curriculum Learning):} We utilize the simple architecture (Config A) but modify the data sampling distribution over time. We employ a stepped curriculum schedule: for the first 25\% of training steps ($t < 500$), values are sampled from a small range $x \in [0, 100]$. The range is expanded to intermediate complexity ($x \in [0, 1000]$) for the next 25\% ($t < 1000$), and finally to the full Out-of-Distribution range ($x \in [0, 5000]$) for the remaining 50\%. \textit{Hypothesis Tested:} Is the failure driven by complexity overload? This tests if the ``Linear Trap'' is a result of the model being overwhelmed by the high variance of the full distribution early in training. By restricting the initial training to small integers, we test if the model can first learn the logic in a low-noise regime (where linear shortcuts are less dominant) and then robustly transfer that logic to larger magnitudes. \textit{Structural Note:} The failure of this configuration (as observed in results) is particularly insightful. It suggests that the model ``locks in'' to the linear solution during the easy phase (where $y \approx x$ is a very strong proxy) and cannot unlearn it even when the data distribution shifts, highlighting the path-dependence of the optimization trap.

\end{itemize}

\textbf{Group B: Topological Constraints}
\begin{itemize}
    \item \textbf{Config B$_{fl}$ (Frozen Logic):} We initialize and freeze the FFN and Readout weights to the exact analytical solution of the ``Geometric ALU'' (see Appx.~\ref{app:gALU}), effectively creating a fixed, differentiable program. Only the Attention layers ($W_Q, W_K, W_V$) remain trainable. \textit{Hypothesis Tested:} Is the failure located in the Logic (FFN) or the Routing (Attention)? By freezing the downstream logic to a known-perfect state, we test if the optimizer can solve the ``Binding Problem'' (learning to attend to operators). Failure here confirms the ``Gradient Wall'' hypothesis: it indicates that even with a perfect functional destination, the gradients generated by discrete, saturated logic gates ($\sigma(x) \approx 0 \text{ or } 1$) vanish before they can update the upstream attention mechanism. \textit{Structural Note:} The frozen logic implements the standard ``Cut-and-Paste'' subtraction algorithm ($y = D_0 - 2D_5$) triggered by an AND gate on the Type flags ($IsNum \land IsSub$). We use high-gain weights ($w=100$) to ensure the gates are stiff, mimicking the conditions of a converged binary circuit.

    \item \textbf{Config B$_{fa}$ (Perfect Routing):} We utilize \textbf{Frozen Oracle Attention}, fixing the attention weights to perform perfect token routing (Numbers attend to Operators). The raw attention output ($z \in [0,1]$) is added directly to the high-magnitude residual stream ($x \in [0, 5000]$). \textit{Hypothesis Tested:} Is the ``Binding Problem'' the primary bottleneck? By solving the routing problem perfectly for the model, we test if the optimizer's failure to discover logic is simply a failure to find the correct data associations. A failure here (as observed) isolates \textbf{Data-Control Interference} as a distinct pathology from Attention Failure: even when the model ``knows'' the operator is a Minus sign, the signal is too weak relative to the payload to trigger the logic gate. \textit{Structural Note:} This configuration serves as the direct ablation for the Signal Boosting experiments (Group C). It establishes the baseline performance of a ``correctly wired'' but ``energetically imbalanced'' network.
    
    \item \textbf{Config B$_{ortho}$ (Value-Orthogonality):} We utilize the simple architecture of Config A but augment the optimization objective with a targeted \textbf{Value-Independence Penalty}. We add a regularization term $\mathcal{L}_{reg} = \lambda \sum (W_{gate}[:, D_0])^2$ (with $\lambda=10.0$) to the loss. This specifically penalizes the weights connecting the Value Stratum ($D_0$) to the Logic Gate, forcing the gate to be orthogonal to the numerical payload. \textit{Hypothesis Tested:} Can stratification be induced via soft constraints? This configuration tests if the ``Simplicity Bias'' (the Linear Trap) can be overridden by explicitly punishing the specific linear shortcut the model prefers. If the optimizer is discouraged from using the high-magnitude Value channel for decision-making, we test if it naturally falls back on the Control channels ($D_1-D_5$) or if it fails to converge entirely. \textit{Structural Note:} This represents a ``Soft Stratification'' approach. Unlike the final STRAT architecture which physically removes these connections (Hard Stratification), this configuration leaves them topologically available but energetically expensive, testing the plasticity of the optimization landscape.
    
\end{itemize}

\textbf{Group C: Signal Injection}
\begin{itemize}
    \item \textbf{Config C$_{sb}$ (Signal boost):} We introduce a fixed, scalar pre-activation gain $\gamma=100$ applied to the output of the Attention block, immediately preceding the residual connection ($x_{res} = x + \gamma \cdot \text{Attention}(x)$). All other parameters remain learnable. \textit{Hypothesis Tested:} Does the ``Gradient Wall'' stem from signal attenuation? Standard Dot-Product Attention (softmax) often yields diffuse probability distributions (e.g., $p \approx 0.2$), resulting in weak updates that fail to drive the subsequent sigmoid gates ($\sigma(x)$) into their saturated, non-linear regime. By injecting a static gain $\gamma=100$, we force the attention signal to be magnitude-compatible with the stiff logical thresholds required by the FFN, testing if this simple scaling resolves the optimization deadlock. \textit{Structural Note:} The gain is applied \textit{post-projection} but \textit{pre-residual}. This is crucial: it amplifies the ``update'' signal (the Type information) relative to the pass-through ``residual'' (the Value payload), effectively altering the signal-to-noise ratio in favor of the control logic without altering the FFN architecture itself.
    
    \item \textbf{Config C$_g$ (Learned Gain):} We introduce a single \textbf{learnable scalar parameter} $\alpha$, initialized to $1.0$, which scales the output of the Attention block before the residual connection ($x_{res} = x + \alpha \cdot \text{Attention}(x)$). \textit{Hypothesis Tested:} Can the optimizer autonomously discover the high-gain regime required for logic? This tests the ``Chicken-and-Egg'' dilemma of the Gradient Wall: to get strong gradients from the sigmoid gate, the input signal must be large (high gain); but to learn a large gain, the model needs strong gradients. Initializing $\alpha=1.0$ tests if there is a viable path across the optimization landscape from the standard operating point to the high-gain solution. \textit{Structural Note:} We use a scalar parameter rather than a vector to prevent the model from using the gain term to fundamentally alter the direction of the attention update (i.e., reshaping the embedding). It acts strictly as a \textbf{volume knob} for the Control Signal.

    \item \textbf{Config C$_b$ (Balanced Operating Point):} We utilize \textbf{Frozen Oracle Attention} (Config B$_{fa}$), initializing and fixing the attention weights to perform perfect token routing (see Appx.~\ref{app:gALU}). To address the signal-to-noise imbalance, we initialize a learnable gain parameter to $\alpha=50.0$. This ensures the Control signal entering the residual stream is of comparable magnitude to the Data payload. \textit{Hypothesis Tested:} Does a valid \textit{Operating Point} exist? By manually placing the model in a regime with high Signal-to-Noise Ratio (SNR) and perfect routing, we test if the failure to learn is due to the impossibility of the task or simply the difficulty of reaching this basin of attraction. We essentially ``teleport'' the optimizer to the vicinity of a good solution to see if it can converge. \textit{Structural Note:} Unlike Config B$_{fl}$ (Frozen Logic), here we freeze the \textit{Routing} and learn the \textit{Logic}. The initialization of $\alpha=50.0$ acts as a manual ``balancing'' of the subspace variances, preventing the Data Stratum from drowning out the Control Stratum during the initial critical gradients.

    \item \textbf{Config C$_c$ (Synergistic Constraints):} We combine the structural intervention of \textbf{Signal Injection} (C$_{sb}$) with the optimization constraint of \textbf{Soft Stratification} (B$_{ortho}$). The model utilizes a learnable pre-activation gain initialized to $\alpha=50.0$ to ensure the Control signal is visible (addressing the Gradient Wall). Simultaneously, we apply the Value-Independence Penalty ($\lambda=10.0$) to the loss function to actively discourage the linear shortcut (addressing the Linear Trap). \textit{Hypothesis Tested:} Can we engineer a viable optimization path by attacking both failure modes simultaneously? This configuration tests the ``Carrot and Stick'' hypothesis: the gain $\alpha=50$ acts as the ``Carrot'' (making the logic gate accessible), while the penalty $\lambda$ acts as the ``Stick'' (making the linear shortcut expensive). We test if this dual pressure is sufficient to guide the optimizer into the correct topological solution. \textit{Structural Note:} While Config C$_{sb}$ uses a fixed gain, here we employ a \textbf{learnable parameter} initialized to the high-gain regime ($\alpha=50$). This provides the necessary initial signal strength to overcome the Sigmoid Tax, while allowing the model to fine-tune the gain magnitude during convergence.
    
    \item \textbf{Config C$_{bi}$ (Bias Initialization):} This is identical to Config A, but we modify the initialization of the FFN Logic Gates by explicitly setting the bias terms to $b_{gate} = +1.0$ (replacing the standard uniform initialization). This sets the initial activation of the sigmoid gates to $\sigma(1.0) \approx 0.73$. \textit{Hypothesis Tested:} Is the failure caused by ``Dead Gates''? Standard initialization can sometimes place logic gates in a regime where they block flow or provide weak gradients, causing the optimizer to abandon the path. By forcing the gates into a conductive (``Open'') state at initialization, we test if maintaining strong initial gradient flow allows the model to eventually learn the discrete closing mechanism, or if it traps the model in a linear local minimum. \textit{Structural Note:} This initialization places the model in a \textbf{Pseudo-Linear Regime}. Since the gate activation is roughly constant ($\approx 0.73$) and conductive at start, the GLU behaves like a linear projection ($y \approx 0.73 W_{val}x$). This lowers the barrier to learning arithmetic but raises the risk of the ``Open Gate Trap,'' where the model overfits the values without ever learning to shut the gate (switch) for distractor tokens.

\end{itemize}

\subsection{Detailed Failure Analysis}
\label{app:failure_analysis}
\subsubsection{The ``Honest'' Failures}
\vspace{-0.1in}
High ID error reveals that these models failed to learn the logic gate entirely, defaulting to heuristic baselines.

\textbf{The Linear Trap (Configs A, A$_s$, A$_{curr}$, B$_{ortho}$):} 
The Baseline (\textbf{Config A}) achieves a median ID MAE of $15.5$, suggesting it is merely tracking the input mean or applying a linear regression on the value channel ($y \approx x$). 

Regarding scale, \textbf{Config A$_s$ (Scaled Inputs)} also failed to resolve the problem. In fact, scaling the inputs to $[0,1]$ resulted in worse performance (ID MAE $98.7$), likely because normalizing the massive value channel to the same magnitude as the sparse control flags deprived the optimizer of the only strong gradient signal it had.

Similarly, \textbf{Config A$_{curr}$ (Curriculum)} yielded the highest errors of all (ID MAE $600.3$). This result indicates that the model ``locked in'' to a linear solution during the early, easy phase ($x \in [0,100]$) and could not unlearn it when the distribution shifted, highlighting the path-dependence of the trap.

\textbf{The Gradient Wall (Configs B$_{fl}$, C$_{sb}$, B$_c$):} 
Medians in the range of $\approx 45\text{-}60$ (e.g., \textbf{Config B$_{fl}$} at $58.7$) indicate models stuck between the Linear Trap and a random baseline. This validates the ``Gradient Wall'' hypothesis: even when we freeze the logic to be perfect (B$_{fl}$) or boost the signal (C$_{sb}$), the gradients vanish at the saturation points of the sigmoid gates, stranding the weights in a local minimum where the gate remains permanently closed or open.

\subsubsection{The ``Dishonest'' Successes}
\vspace{-0.1in}
A subset of configurations appeared successful during training (low ID error) but failed catastrophically on OOD data. This is the most dangerous failure mode, as it masquerades as a solution.

\textbf{Magnitude Leaks (Config B$_{fa}$ \& C$_b$):} 
\textbf{Config B$_{fa}$} (Frozen Oracle Attention) achieved an excellent ID MAE of $6.2$, implying it had solved the task. However, its OOD error exploded to \textbf{11,817}. Even with perfect routing provided by the oracle, the dense layers used the \textit{value magnitude} as a proxy for the decision boundary rather than the boolean flag. 
Similarly, \textbf{Config C$_b$} (Balanced Operating Point) achieved the best training score of the entire ablation (ID MAE $1.8$) but the worst generalization (OOD MAE $11,916$). This proves that fixing the signal-to-noise ratio via initialization merely allows the model to overfit the operating point more efficiently, without discovering the topological switch.

\textbf{The Open Gate Trap (Config C$_{bi}$):} 
Initializing gate biases to $+1.0$ allowed the model to achieve a respectable ID MAE of $12.7$. However, the OOD error ($8,471$) confirms that the gates simply remained open, acting as a linear pass-through that failed to block distractor tokens in the high-magnitude regime.

% \newpage
%%%%%%%%%%%%%%%%%%%%%%%%%%%%%%%%%%%%%%%%%%%%%%%%%%%%%%%%%%%%%
%% Appendix C
%%%%%%%%%%%%%%%%%%%%%%%%%%%%%%%%%%%%%%%%%%%%%%%%%%%%%%%%%%%%%
\section{The Two Regimes of Learned Logic}
\label{app:alien_logic}

Our analysis of the learned weights across 10 independent training runs reveals that while the STRAT-12 architecture consistently converges to the same topological mechanism (the ``Cut-and-Paste'' router), the specific implementation depends on the sign initialization of the Readout Layer. This symmetry breaking results in two distinct logical regimes: the \textbf{Standard Attractor} (Add-by-Default) and the \textbf{Mirror Attractor} (Subtract-by-Default).

Here we present the raw weights from two representative runs to demonstrate these algorithms. These weights were extracted after 2,000 steps of training.

\subsection{Run 1: The Standard Attractor (Seed 42)}

\begin{verbatim}
[1] READOUT LAYER (W_out)
Equation: y = sum(W_out * x)
        D0(Val)    D1(Num)     D2(+)     D3(-)    D4(Op)  D5(Scratch)
Output   1.0005    -0.0149   -0.0123   -0.2240    0.2774       1.1755

[2] THE ROUTER (FFN)
(A) Logic Gates (W_gate)
Input Columns: D1(Num), D2(+), D3(-), D4(Op)
           D1(Num)     D2(+)     D3(-)    D4(Op)       Bias
Gate_D0    -0.7223   -2.7710    2.8255    2.6739     0.0483
Gate_D5    -0.5233   -3.3227    2.4424    2.5473    -0.1033

(B) Value Routing (W_val)
Input Columns: From_D0, From_D5
           From_D0   From_D5
To_D0      -1.0120   -0.0112  (The "Cut": Adds ~ -1.0x to D0)
To_D5      -0.8460    0.0017  (The "Paste": Adds ~ -0.85x to D5)

[3] ATTENTION LAYER (Control Subspace Only)
Rows/Cols correspond to flags: D1(Num), D2(+), D3(-), D4(Op)

(A) Query (W_q)
          D1       D2       D3       D4
D1    0.4299  -0.8417  -1.4502  -1.3617
D2    0.3148  -0.9022  -1.1602  -1.1371
D3    0.8778  -1.7965  -0.1897  -1.1168
D4    0.9766  -1.0427  -0.0547  -0.7767

(B) Key (W_k)
          D1       D2       D3       D4
D1   -0.2308  -0.5511   0.5377   0.7437
D2   -0.6876  -0.9046   0.4638   0.7889
D3   -0.7544  -2.5763   0.3965   0.0591
D4   -0.2349  -2.4472   0.9284  -0.1426

(C) Value (W_v)
          D1       D2       D3       D4
D1    0.8222   0.1646  -0.2627   0.0101
D2    1.6802   0.8191  -1.0734  -1.0369
D3   -1.3298  -0.2486   0.9975   1.2483
D4   -1.2396  -0.7179   1.3799   0.9310
\end{verbatim}

\textbf{Overall Interpretation:} Standard logic. Positive Readout. The Attention layer (W\_v) passes flags largely unmodified (positive weights on diagonal), and the Gate opens on MINUS ($D_3$) to trigger the Cut-and-Paste.

\textbf{Condition:} Readout Layer initializes with positive weights ($w_{out} > 0$).
\textbf{Strategy:} The model treats the residual stream as an ``Addition Conveyor Belt.'' It performs addition lazily (by doing nothing) and subtraction actively (by destroying the payload and creating a negative copy).

\paragraph{1. The Readout Layer}
The readout weights for the input ($D_0$) and scratchpad ($D_5$) are positive.
\begin{equation}
    y \approx 1.00 \cdot D_0 + 1.18 \cdot D_5
\end{equation}
\textit{Implication:} If the logic gates stay closed ($D_0=x, D_5=0$), the output is $+x$.

\paragraph{2. The Gate Logic}
The gates are biased to be closed ($\approx 0$), and learn positive weights for the \texttt{MINUS} flag ($D_3$).
\begin{align*}
    \text{Bias}_{D0} &\approx 0.05 \\
    w_{gate}(D_3) &\approx +2.83 \quad (\text{Opens on Minus}) \\
    w_{gate}(D_2) &\approx -2.77 \quad (\text{Closes on Plus})
\end{align*}

\paragraph{3. The Active Routing (Minus Token)}
When triggered, the Value Router performs the ``Cut-and-Paste'' operation:
\begin{itemize}
    \item \textbf{Cut ($D_0 \to D_0$):} The update is $-1.01x$, canceling the residual $x$. ($D_0 \to 0$).
    \item \textbf{Paste ($D_0 \to D_5$):} $w \approx -0.85$. A negative copy is moved to $D_5$. ($D_5 \to -0.85x$).
\end{itemize}
\textbf{Total Output:} $1.0(0) + 1.18(-0.85x) \approx \mathbf{-x}$.

\subsection{Run 2: The Mirror Attractor (Seed 1000)}

\begin{verbatim}
[1] READOUT LAYER (W_out)
Equation: y = sum(W_out * x)
        D0(Val)    D1(Num)     D2(+)     D3(-)    D4(Op)  D5(Scratch)
Output  -1.0025     0.1690    0.5371    0.0803    0.6230      -1.0771

[2] THE ROUTER (FFN)
(A) Logic Gates (W_gate)
Input Columns: D1(Num), D2(+), D3(-), D4(Op)
           D1(Num)     D2(+)     D3(-)    D4(Op)       Bias
Gate_D0     1.6707   -2.4147    1.8345    1.6123     0.8002
Gate_D5     1.5939   -2.1411    1.8549    1.7351     0.7962

(B) Value Routing (W_val)
Input Columns: From_D0, From_D5
           From_D0   From_D5
To_D0      -0.8788    0.0005
To_D5      -1.0361   -0.0021

[3] ATTENTION LAYER (Control Subspace Only)
Rows/Cols correspond to flags: D1(Num), D2(+), D3(-), D4(Op)

(A) Query (W_q)
          D1       D2       D3       D4
D1   -1.2564  -0.6667   0.2862  -0.6111
D2   -0.7708   0.4338  -1.2994  -0.4427
D3    1.0786   0.7103   0.5763   0.7327
D4   -0.8597  -0.4932  -0.6717  -0.9511

(B) Key (W_k)
          D1       D2       D3       D4
D1    0.9547  -1.3567  -0.3891  -1.2468
D2    0.2952  -0.3929  -1.2440  -1.1796
D3   -1.1966   0.9828   0.8603   1.0142
D4    0.7991  -0.5533  -0.7871  -1.2244

(C) Value (W_v) -- Note the inversion on D2(+) and D3(-)
          D1       D2       D3       D4
D1   -0.0692   1.2879  -1.2549  -1.0564
D2   -0.8868  -2.3064   1.0775   0.6756  <-- CRITICAL INVERSION
D3    0.4572   0.8875  -0.7919  -0.0625
D4    0.8125   1.2767  -0.2441  -1.2211
\end{verbatim}

\textbf{Overall Interpretation:} Inverted logic. Negative Readout. Note the critical inversion in \texttt{W\_v} (Row D2, Col D2 is -2.30), which flips the PLUS flag to trigger the ``Cut-and-Paste'' logic, effectively turning addition into a double-negative operation.

\textbf{Condition:} Readout Layer initializes with negative weights ($w_{out} < 0$).
\textbf{Strategy:} The model treats the residual stream as a ``Subtraction Conveyor Belt.'' It performs subtraction lazily (by doing nothing) and addition actively (by inverting the logic to force a double-negative).

\paragraph{1. The Readout Layer}
The readout weights are inverted.
\begin{equation}
    y \approx -1.00 \cdot D_0 - 1.08 \cdot D_5
\end{equation}
\textit{Implication:} If the logic gates stay closed ($D_0=x, D_5=0$), the output is $-1.00(x) = \mathbf{-x}$. The default state is Subtraction.

\paragraph{2. The Logic Inversion}
To perform Addition, the model must trigger the ``Cut-and-Paste'' mechanism on \texttt{PLUS} tokens. However, the gate weights initially appear paradoxical:
\begin{align*}
    w_{gate}(D_2 \text{ aka Plus}) &\approx -2.41 \quad (\text{Naively Closes on Plus}) \\
    w_{gate}(D_3 \text{ aka Minus}) &\approx +1.83 \quad (\text{Naively Opens on Minus})
\end{align*}
Under standard flags, this would fail (Closing on Plus would yield the default $-x$). The model resolves this by \textbf{inverting the control flags in the Attention Layer}.
\begin{itemize}
    \item The Attention matrix $W_V$ learns a large negative weight ($-2.31$) for the self-attention of the \texttt{PLUS} flag.
    \item A \texttt{PLUS} token ($1.0$) is transformed into a negative flag ($\approx -1.3$).
    \item \textbf{Gate Activation:} $(-1.3) \times (-2.41) \approx \mathbf{+3.13}$.
\end{itemize}
The gate effectively \textbf{Opens on Plus} due to the double inversion (Negative Flag $\times$ Negative Weight).

\paragraph{3. The Active Routing (Plus Token)}
When triggered, the router executes the same geometric move:
\begin{itemize}
    \item \textbf{Cut ($D_0 \to D_0$):} $w \approx -0.88$. ($D_0 \to \approx 0$).
    \item \textbf{Paste ($D_0 \to D_5$):} $w \approx -1.04$. ($D_5 \to -1.04x$).
\end{itemize}
\textbf{Total Output:}
Because the readout is negative, the pasted negative value becomes positive:
\begin{equation}
    y \approx -1.08 \cdot D_5 = -1.08 \cdot (-1.04x) \approx \mathbf{+x}
\end{equation}

\subsection{Synthesis}
Both regimes implement an identical geometric operation: an orthogonal basis exchange enabled by the topological isolation of the magnitude subspace.

\begin{table}[ht]
\centering
\caption{Comparison of the two learned algorithms.}
\begin{tabular}{l c c}
\toprule
\midrule
\textbf{Feature} & \textbf{Standard (Run 1)} & \textbf{Mirror (Run 2)} \\
\midrule
\textbf{Readout Sign} & Positive ($+$) & Negative ($-$) \\
\textbf{Default Operation} & Addition ($+x$) & Subtraction ($-x$) \\
\textbf{Active Trigger} & \texttt{MINUS} Token & \texttt{PLUS} Token \\
\textbf{Attention Role} & Pass-through & Signal Inversion \\
\textbf{Resulting Logic} & $y = \text{Default} - \text{Routed}$ & $y = \text{Default} + \text{Routed}$ \\
\midrule
\bottomrule
\end{tabular}
\end{table}

%%%%%%%%%%%%%%%%%%%%%%%%%%%%%%%%%%%%%%%%%%%%%%%%%%%%%%%%%%%%%
%% Appendix D
%%%%%%%%%%%%%%%%%%%%%%%%%%%%%%%%%%%%%%%%%%%%%%%%%%%%%%%%%%%%%
\section{Evaluation Across Diverse Tasks: Experimental Details}
\label{app:dataset_configurations}%
\label{app:model_configurations}%
\label{app:tasks_and_datasets}%

This appendix documents the 11 evaluations reported in Section~\ref{sec:diverse_tasks}
(Table~\ref{tab:task_overview}). Each section covers one dataset in a fixed order: the source data and prediction target, the construction of the training, test, and OOD test sets, the model configuration, and the training configuration.

\begin{table}[htbp]
\centering
\captionsetup{font=footnotesize}
\caption{Trainable parameter counts for STRAT and the Transformer baseline across the 11 dataset evaluations.}
\label{tab:parameter_counts}
{\small
\begin{tabular}{lrr}
\toprule
\midrule
\textbf{Dataset} & \textbf{STRAT parameters} & \textbf{Transformer parameters} \\
\midrule
SATNet Sudoku & 396,335 & 805,349 \\
SATBench & 1,648 & 5,310 \\
SATLIB & 1,648 & 5,310 \\
Tic-Tac-Toe Endgame & 33,474 & 58,498 \\
Game of Life & 17,993 & 30,761 \\
PMLB Parity5 & 35,314 & 60,594 \\
PMLB M-of-N & 17,722 & 30,426 \\
PMLB MONK-2 & 17,722 & 30,426 \\
PMLB LED24 & 17,722 & 30,426 \\
PMLB ThreeOf9 & 17,722 & 30,426 \\
SPECT Heart & 17,722 & 30,426 \\
\midrule
\bottomrule
\end{tabular}
}
\end{table}

\subsection{SATNet Sudoku}
SATNet Sudoku provides 1,100 pairs of $9\times9$ puzzles and completed solutions \citep{wang2019satnet}, split into 100 training and 1,000 test puzzles. The task fills every empty cell while preserving the given digits and Sudoku constraints.

\subsubsection{Dataset Construction}
\textbf{Training-set construction.}
We use the first 10 puzzles from the 100-puzzle training split as the base training set. At each epoch, we replace each base puzzle with a randomly transformed version and apply the same transformation to its solution. The transformation randomly relabels the nine digits, permutes the three row bands and the rows within each band, applies the corresponding permutations to column stacks and columns, and transposes the grid with probability 0.5. 

\textbf{Test-set construction.}
We combine the other 90 puzzles from that split with the original 1,000 test puzzles, yielding 1,090 test boards.

\textbf{OOD test-set construction.}
A \emph{band} is a horizontal group of three rows, and a \emph{stack} is a vertical group of three columns. For each puzzle, we define $I$ as the maximum of two quantities: the range of clue counts across the three bands and the range across the three stacks. Across the 1,100 source puzzles, the mean clue count is 36.21 and the mean $I$ is 3.10. In total, 97.1\% of the puzzles satisfy $I\leq5$, and the largest $I$ among the 10 training puzzles is 5. From the remaining 1,090 puzzles, we select all eight puzzles with exactly 35 clues and $I\in\{6,7\}$. We then apply the same Sudoku automorphisms used for training augmentation to generate 500 unique puzzle--solution pairs. Because these automorphisms preserve $I$, every generated puzzle retains $I\in\{6,7\}$. The resulting OOD set therefore lies outside the support of the augmented training set with respect to $I$. For consistent presentation and evaluation orientation, we rotate each pair so that its densest band occupies the bottom three rows.

\subsubsection{Model Configuration}
We represent each of the 81 cells with one token. STRAT assigns 64 channels to Type and 18 to Data. Thirty-seven Type channels encode row, column, and box membership, whether the cell is given, and the given digit. The other 27 channels start at zero. Because a given digit fixes the constraints on its cell, we treat it as immutable Type information. Data allocates nine channels to a digit register and nine to a candidate register. Given cells initialize the digit register with a one-hot digit; every other Data channel starts at zero.

STRAT stacks 16 four-head attention layers. It computes queries and keys from Type, then uses the resulting routes to carry Data into a width-128 GLU with
one hidden layer. Separate LayerNorm modules normalize Type and Data. A final Data readout produces nine digit logits per cell. 

The Transformer concatenates the same 64 Type channels and 18 Data channels, then adds two zero channels to form a width-84 stream divisible across four heads. Its 16 attention layers use a one-hidden-layer FFN of width 128. Self-attention and the FFN update the full stream, and the readout produces nine digit logits per cell.

\subsubsection{Training Configuration}
We train both models on the augmented base puzzles for 6,000 epochs using AdamW with a learning rate of $10^{-3}$. The loss is cross-entropy over the empty cells, and gradients are clipped to a norm of 1.0. For each paired seed from 41 to 50, we evaluate the checkpoint from the final epoch. We count a board as correct only if all 81 cells match the reference solution, with the given cells copied directly from the input puzzle.

\subsection{SATBench}
SATBench contains 2,100 Boolean formulas in conjunctive normal form (CNF), derived from natural-language logic scenarios \citep{wei2025satbench}. We use these formulas to construct an assignment-verification task. Given a formula and a Boolean assignment, predict whether the assignment satisfies every clause.

\subsubsection{Dataset Construction}
\textbf{Training-set construction.}
The dedicated training split contains 10 satisfiable formulas with 4--50 clauses and 16--90 variables. A SAT solver provides one satisfying assignment for each formula. At every epoch, we resample training examples from these 10 formula--assignment pairs.  Each epoch contains 16 positive and 16 negative examples. For a positive example, we use the solver assignment directly. To construct a negative example, we perturb the solver assignment by flipping one, two, or three variables and select the first perturbation that violates at least one clause. If none of these perturbations produces a violation, we flip five variables instead. 

\textbf{Test set.}
The test split contains 2,090 formulas with 4--50 clauses: 1,040 satisfiable and 1,050 unsatisfiable. Each satisfiable formula is paired with a solver-produced satisfying assignment and labeled positive, while each unsatisfiable formula is paired with a seeded random assignment and labeled negative.

\textbf{OOD test set.}
From the test set, we sample 250 positive and 250 negative examples and cyclically repeat each formula's clause sequence until it contains exactly 100 clauses.

\subsubsection{Model Configuration}
We serialize each formula with eight token types: evaluated true literal, evaluated false literal, OR, AND, clause start, clause end, end of sequence, and padding. An eight-dimensional one-hot vector serves as the Type state. Data has three channels. The first holds $+1$ for a true literal, $-1$ for a false literal, and zero for structural tokens; the other two provide workspace.

Four single-head attention layers process the sequence. In STRAT, attention reads and updates the eight-dimensional Type state, while a width-32 GLU with one hidden layer updates the width-3 Data state.

The Transformer packs the same fixed fields into an 11-dimensional residual stream. It has four single-head layers and a width-32 FFN with one hidden layer. Attention and the FFN update the full stream.

\subsubsection{Training Configuration}
Each epoch draws a balanced batch from the 10 training formulas: 16 satisfying assignments and 16 violating assignments. A SAT solver supplies the satisfying assignments. We search for violating assignments by flipping one to three variables; if that search fails, the implementation applies a five-variable fallback flip. STRAT and the Transformer receive the same sampled formula and assignment pairs. We train for 2,000 epochs with Adam at learning rate $10^{-2}$, binary cross-entropy, and gradient clipping at norm 1.0. The final-epoch checkpoint yields assignment-level accuracy for each paired seed in 41--50.

\subsection{SATLIB}
The SATLIB collection we use contains 2,000 random 3-CNF formulas over 20 variables \citep{hoos2000satlib}: 1,000 satisfiable formulas with 91 clauses and 1,000 unsatisfiable formulas with 110 clauses. Every clause has exactly three literals. SATLIB labels each formula as satisfiable or unsatisfiable; we instead use the formulas for the same assignment-verification task as SATBench, with the same definition of a violated clause.

\subsubsection{Dataset Construction}
\textbf{Training set.}
The first 10 satisfiable formulas in file order form the training pool. We generate training examples using the same procedure as in SATBench. At each epoch, we sample 16 positive examples from solver-produced satisfying assignments and 16 negative examples by flipping a small number of variables in those assignments.

\textbf{Test set.}
The remaining 1,990 formulas form the ID test set. For the 990 satisfiable formulas, positive examples use solver-produced satisfying assignments. For the 1,000 unsatisfiable formulas, negative examples use random assignments drawn from the evaluation seed.

\textbf{OOD test set.}
We select the first 250 satisfiable formulas from the test set in file order. For each formula, we use its solver-produced satisfying assignment as a positive example and create a negative example by flipping one variable. This produces 500 examples over 250 formulas with balanced labels. We then append the first nine clauses to each 91-clause formula, producing a 100-clause formula.

\subsubsection{Model Configuration}
SATLIB uses the SATBench model configuration and the same eight token types. STRAT has an eight-dimensional Type state, a width-3 Data state, four single-head layers, and a width-32 GLU with one hidden layer. 

The Transformer packs the same fixed fields into an 11-dimensional stream. Its four single-head layers and a width-32 FFN with one hidden layer.

\subsubsection{Training Configuration}
Each epoch draws 16 satisfying and 16 violating assignments from the 10 training formulas, following the SATBench construction. We train both models for 2,000 epochs with Adam at learning rate $10^{-2}$, binary cross-entropy, and gradient clipping at norm 1.0. For each paired seed in 41--50, the final checkpoint gives assignment-level accuracy on the 1,990 test formulas.

\subsection{Tic-Tac-Toe Endgame}
The UCI Tic-Tac-Toe Endgame dataset contains 958 legal terminal board positions \citep{tic-tac-toe_endgame_101}. Each board represents its nine cells as \emph{X}, \emph{O}, or blank. We retain the original binary label, which is positive when \emph{X} forms a complete row, column, or diagonal.

\subsubsection{Dataset Construction}
\textbf{Training set.}
We select 10 base boards from the 958 examples. Because the winning-line rule is invariant to rotations and reflections, we apply all eight symmetries of the square to each base board and remove duplicates. This produces 76 unique training boards, which we use to train both models.

\textbf{Test set.}
The remaining 948 boards form the initial test set. Among them, 66 are rotations or reflections of a base board and therefore appear in the augmented training set. We exclude these 66 boards, leaving 882 boards for the ID evaluation.

\textbf{OOD test set.}
Ten base boards cannot cover every winning configuration, so the augmented training set contains incidental correlations between mark locations and the label. We construct the OOD set from test boards whose spatial statistics favor the opposite label under these correlations. For each board, we count \emph{X}, \emph{O}, and blank cells within three regions: the center, the four corners, and the four edges. These counts form a nine-dimensional vector $r(x)$ that is invariant to rotations and reflections. Let $c$ be the difference between the mean $r(x)$ of positive and negative training boards. A positive score $r(x)^\top c$ indicates greater similarity to the positive training boards under these spatial statistics.
From the 948 test boards, we retain positive boards with $r(x)^\top c<0$ and negative boards with $r(x)^\top c>0$, yielding 216 positive and 119 negative candidates. Removing boards that appear in the augmented training set leaves 195 positive and 112 negative boards. We downsample the positive candidates to 112 boards. The resulting OOD test set contains 112 positive and 112 negative boards.

\subsubsection{Model Configuration}
We encode each board as one classification token, nine cell tokens, and eight line tokens. The line tokens represent the three rows, three columns, and two diagonals. STRAT represents every token with separate width-32 Type and Data streams. For a cell token, Type encodes its role, row, column, diagonal membership, and immutable board value. Line-token Type identifies the line and its member cells. The board value belongs to Type because it determines how the corresponding cell participates in each candidate line.

Each cell token initializes four identical eight-channel Data registers. A register contains a seven-way categorical field and a presence bit. The classification and line tokens start with zero Data.

STRAT applies two four-head attention layers and a two-hidden-layer GLU with hidden widths 64 and 64. The Transformer concatenates the same observable Type and Data fields into one width-64 residual stream. It applies two four-head attention layers and a two-hidden-layer FFN with hidden widths 64 and 64. Both models predict from the classification token.

\subsubsection{Training Configuration}
We apply the eight rotations and reflections of the square to each base board, then remove duplicate configurations. Both models receive the same augmented set. Training runs for 2,000 epochs with full-batch cross-entropy, AdamW at learning rate $10^{-3}$, and gradient clipping at norm 1.0. For paired seeds 41--50, the final-epoch checkpoint provides board-level classification accuracy.

\subsection{Game of Life}
From the public \emph{gameoflife-v1} collection, we retain 767 one-step transitions on $6\times6$ boards.\footnote{\url{https://huggingface.co/datasets/neoneye/gameoflife-v1}} The boards wrap around at the boundaries, giving every cell exactly eight neighbors (a toroidal Moore neighborhood). The target is the board after one update under the B3/S23 rule \citep{strandgaard2023gameoflife}. Under this rule, a dead cell becomes alive when exactly three neighbors are alive, while a live cell survives when two or three neighbors are alive.

\subsubsection{Dataset Construction}
\textbf{Training set.}
The first 10 board--successor pairs in file order form the base training set. At each epoch, we sample one random transformation and apply it jointly to every board and its successor in the batch. The transformation combines a toroidal translation, a rotation by a multiple of $90^\circ$, a row reflection, and a column reflection. Because the B3/S23 update commutes with these transformations, every transformed pair remains a valid one-step transition.

\textbf{Test set.}
The remaining 757 pairs form the ID test set.

\textbf{OOD test set.}
The B3/S23 update depends on each cell's number of live neighbors. We therefore test whether the models generalize to spatial arrangements that do not occur in training. For each board, we record the number of live cells and the number of live--live adjacencies. A live--live adjacency is an unordered pair of neighboring live cells on the torus. Both quantities are invariant under the training transformations.
The 10 training boards contain 7--32 live cells and at least 10 live--live adjacencies. We generate 500 unique boards with exactly seven live cells: 250 have eight live--live adjacencies and 250 have nine. We compute each successor using the B3/S23 rule. Only 7 of the 757 ID test boards fall in this regime, and none of the 500 generated boards appears in the source collection. Every OOD successor contains at least one live cell, so an all-dead prediction achieves zero exact-match accuracy.

\subsubsection{Model Configuration}
We represent each of the 36 cells with one token. STRAT assigns 32 channels to Type and 32 to Data. Type stores the immutable alive/dead input state as a two-channel one-hot vector (channel 0 dead, channel 1 alive); all other Type channels start at zero. The input bit belongs to Type because it determines which branch of the transition rule governs the cell. 
Data holds the alive bit in channel 0 and initializes the other channels to zero as scratch. 
One four-head STRAT attention layer computes routes from Type and transports Data to a two-hidden-layer GLU with hidden widths 68 and 68.  The Transformer concatenates the same observable fields into a width-64 residual stream. Its single four-head layer uses a two-hidden-layer FFN with hidden widths 68 and 68.

\subsubsection{Training Configuration}
We train both models on the same 10 board pairs for 6,000 epochs. At each epoch, we apply the same randomly sampled toroidal translation, rotation, and row and column reflections to the full batch. Optimization uses AdamW at learning rate $10^{-3}$, binary cross-entropy over all 36 cells, and gradient clipping at norm 1.0. We evaluate the final-epoch checkpoint for paired seeds 41--50. A prediction counts as correct only when all 36 cells match the reference next state.

\subsection{PMLB Parity5}
The Parity5 task uses ten-bit inputs. The first five bits are one of the 32 assignments listed in PMLB Parity5 \citep{olson2017pmlb}; we call them the relevant bits. The other five bits are \emph{nuisance bits} that always share one value, 0 or 1. The label is the parity of the five relevant bits: an input is positive when an odd number of them are active. The nuisance bits never affect the label. The input space contains $32\times2=64$ distinct inputs.

\subsubsection{Dataset Construction}
\textbf{Training set.}
We randomly select 10 of the 32 assignments. For each selected assignment, the training set contains four copies with the nuisance value equal to the parity label and one copy with the opposite nuisance value. This produces 50 training examples over 20 distinct inputs, with the nuisance value agreeing with the label in 80\% of the examples.

\textbf{Test set.}
The test set contains 160 class-balanced examples covering all 64 distinct inputs. Each input whose nuisance value agrees with the label appears four times, whereas each disagreeing input appears once. The nuisance value therefore agrees with the label in 80\% of the examples, matching the training distribution.

\textbf{OOD test set.}
The OOD test set contains the same 64 inputs with the frequencies reversed: each agreeing input appears once, whereas each disagreeing input appears four times. This produces 160 class-balanced examples with 20\% nuisance--label agreement. The ID and OOD sets have identical input support and use the same parity rule; they differ only in the nuisance--label correlation.

\subsubsection{Model Configuration}
we represent each token with the same 32-dimensional Type representation and 32-dimensional Data representation. STRAT processes these representations as separate streams. In Type, the first dimension identifies the classification token, the next two distinguish relevant from nuisance bits, and the following two encode the observed binary value. The classification token activates its own indicator and the relevant-group indicator. Each bit token activates its corresponding group indicator and one of the two value dimensions. The remaining Type dimensions are initialized to zero.
In Data, the first two dimensions encode the observed binary value of each bit token, while the remaining 30 dimensions are initialized to zero and serve as scratch space for intermediate computation. The classification token's Data is initialized entirely to zero.

STRAT applies two four-head attention layer and a two-hidden-layer GLU with hidden widths 68 and 68. The Transformer concatenates the same observable fields into a width-64 residual stream. It uses two four-head layer and a two-hidden-layer FFN with hidden widths 68 and 68.

\subsubsection{Training Configuration}
We use full-batch cross-entropy, AdamW at learning rate $10^{-3}$, and gradient clipping at norm 1.0 for 2,000 epochs. For paired seeds 41--50, we evaluate the final-epoch checkpoint and report accuracy on the 160-example ID and 160-example OOD test sets described above.

\subsection{PMLB M-of-N}
The PMLB M-of-N dataset contains 1,324 examples with 10 binary features \citep{olson2017pmlb}. A label is positive when at least three of seven designated features are active. Three additional features do not affect the label. The notation $(3,7,10)$ denotes the threshold, the number of relevant features, and the total number of features. We refer to the three irrelevant features as nuisance bits.

\subsubsection{Dataset Construction}
\textbf{Training set.}
We randomly select 10 examples from the 1,324-example dataset. We augment each base example with the seven cyclic shifts of its relevant features while leaving the nuisance features unchanged, producing 70 distinct training inputs

\textbf{Test set.}
After excluding every source row whose input appears in the augmented training set, the remaining 1,231 rows form the test set.

\textbf{OOD test set.}
We construct the OOD test set by shifting the relevant-feature-count distribution from negative examples with zero to two active relevant features and positive examples with three to seven in ID to negative examples with exactly two active relevant features and positive examples with exactly three in OOD. We begin with the 1,231 held-out source rows and retain only the examples satisfying these conditions. The resulting 494 source rows correspond to 378 unique inputs after duplicates are merged. 

\subsubsection{Model Configuration}
We encode each input with one classification token followed by 10 feature tokens. For both models, each token is represented by 32 Type dimensions and 32 Data dimensions. STRAT processes these representations as separate streams. 
In Type, the first dimension identifies the classification token, the next two distinguish relevant from nuisance features, the following 10 identify the individual features, and the next two encode the observed binary value. Each feature token activates its corresponding group indicator, feature identity, and one of the two value dimensions. The classification token activates only its own indicator, and the remaining Type dimensions are initialized to zero.
In Data, the first two dimensions encode the observed binary value of each feature token, while the remaining 30 dimensions are initialized to zero and are available as scratch space for intermediate computation. The classification token's Data is initialized entirely to zero.

STRAT applies one four-head attention layer and a two-hidden-layer GLU with hidden widths 68 and 68. The Transformer concatenates the same observable fields into a width-64 stream. Its single four-head attention layer uses a two-hidden-layer FFN with hidden widths 68 and 68. Both models read the prediction from the
classification token.

\subsubsection{Training Configuration}
Both models receive the same full batch and train for 2,000 epochs. For paired seeds 41--50, we report final-checkpoint accuracy.

\subsection{PMLB MONK-2}
PMLB MONK-2 contains 601 examples with six categorical attributes having 3, 3, 2, 3, 4, and 2 possible values, respectively \citep{olson2017pmlb}. For each attribute, the distinguished value is encoded as $1$. We retain the original binary target: an example is positive exactly when two of the six attributes equal $1$.

\subsubsection{Dataset Construction}
\textbf{Training set.}
We randomly select 10 base examples. We augment the 10 base examples with label-preserving transformations. Because the label depends only on how many attributes equal $1$, two transformations preserve the label. The first permutes the non-$1$ values within each attribute. The second permutes attributes with the same cardinality: the three ternary attributes among themselves and the two binary attributes among themselves. Applying all combinations of these transformations to the 10 base rows and removing duplicates yields 212 unique training inputs. 

\textbf{Test set.}
The remaining 591 rows form the initial test set. We exclude the 288 rows whose inputs occur in the augmented training set, leaving 303 input-disjoint rows.

\textbf{OOD test set.}
We construct the OOD test set by shifting the probability mass assigned to inputs with category-count compositions absent from the augmented training set from 18.48\% in ID to 32.70\% in OOD, while preserving the distribution of the number of attributes equal to $1$. Because the label is positive exactly when two attributes equal $1$, this also preserves the class proportions and the labeling rule. Specifically, we merge duplicate inputs in the 303 ID rows to obtain 220 unique inputs, assigning each input an initial probability proportional to its source frequency. For each input, we count how many attributes take each of the four categorical values. Within each group having the same number of attributes equal to $1$, we multiply the probability of inputs whose count pattern is absent from the augmented training set by four and then renormalize the group.

\subsubsection{Model Configuration}
We encode each example with one classification token followed by six attribute tokens. Each token has a width-32 Type stream and a width-32 Data stream. In Type, the classification token activates only the first dimension. For each attribute token, the first dimension is zero, the next six dimensions encode the attribute identity, and the following four encode its observed categorical value. The remaining 21 Type dimensions are zero.
In Data, each attribute token reserves the first seven dimensions for categorical one-hot encoding; only the first four of these dimensions can be nonzero at initialization for MONK-2. The remaining 25 dimensions are initialized to zero as scratch space. The classification token's Data is initialized entirely to zero.

The STRAT model has one four-head attention layer and a two-hidden-layer GLU with hidden widths 68 and 68. For the Transformer, we place the same observable fields in a width-64 residual stream. It uses one four-head layer and a two-hidden-layer FFN with hidden widths 68 and 68. 

\subsubsection{Training Configuration}
We train for 2,000 epochs. For paired seeds 41--50, the final checkpoint provides accuracy on the test examples.

\subsection{PMLB LED24}
PMLB LED24 contains 3,200 examples with 24 binary features derived from a seven-segment display benchmark \citep{olson2017pmlb}. The first seven features correspond to the display segments, and the remaining 17 are random nuisance features. We construct a binary classification task using this input space, assigning a positive label if and only if at least six of the seven segment features are active (i.e., equal to 1).

\subsubsection{Dataset Construction}
\textbf{Training set.}
We select 10 examples, five per class, from the 3,200-example dataset using a greedy feature-coverage procedure with selection seed 3036. We augment each selected example by enumerating all seven cyclic shifts of its segment features and their reversals, each with and without simultaneous complementation of all 17 nuisance features. These transformations preserve the label because they leave the number of active segment features unchanged. After removing duplicates, we use the same augmented training set for both models.

\textbf{Test set.}
The remaining 3,190 examples, labeled by the same rule, form the test set.

\textbf{OOD test set.}
In the original 3,200 examples, between 1 and 15 of the 17 nuisance features are active; neither the all-zero nor the all-one configuration occurs. To construct the OOD test set, we retain positive examples with exactly six active segment features and 15 active nuisance features, and negative examples with exactly five active segment features and zero or one active nuisance feature. These conditions place the segment counts at the decision boundary and shift the nuisance counts to class-conditional extremes outside their ID ranges. We use a fixed hash ordering to select 250 examples per class, producing 500 unique inputs.

\subsubsection{Model Configuration}
We encode each example with one classification token followed by 24 feature tokens. Each token has 64 dimensions: the first 32 form Type, and the remaining 32 form Data. In Type, the first dimension identifies the classification token. For each feature token, one of the next two dimensions indicates whether it is a display-segment or nuisance feature, one of the following 24 dimensions identifies the feature, and one of the next two dimensions encodes its observed binary value. The remaining three Type dimensions are zero. The classification token activates only its own indicator.
In Data, the first two dimensions encode the observed binary value of each feature token, while the remaining 30 dimensions are initialized to zero and serve as scratch space. The classification token's Data is initialized entirely to zero.

STRAT applies one four-head attention layers and a two-hidden-layer GLU with hidden widths 68 and 68. The Transformer concatenates the same observable fields into a width-64 residual stream. It uses one four-head layers and a two-hidden-layer FFN with hidden widths 68 and 68. Both models predict from the classification token.

\subsubsection{Training Configuration}
We augment the first seven fields with cyclic rotations and reflections. For each transformed example, we either retain or jointly complement the 17 irrelevant fields; both operations preserve the density target. Afterward, we remove duplicates and give both models the same augmented set. We train for 2,000 epochs. For paired seeds 41--50, we report final checkpoint accuracy on the remaining 3,190 examples.

\subsection{PMLB ThreeOf9}

PMLB ThreeOf9 contains all 512 assignments to nine binary variables \citep{olson2017pmlb}. We use this complete input table for nine-bit majority classification. Positive labels require at least five active bits. We select 10 training examples. 

\subsubsection{Dataset Construction}
\textbf{Training set.}
We randomly pick 10 assignments.

\textbf{Test set.}
The other 502 assignments form the test set.

\textbf{OOD test set.}
We construct the OOD set by retaining assignments whose bits at positions 4, 6, and 9 contain at least two 1s for a negative label, or at most one 1 for a positive label. 
The true label remains positive when at least five of the nine input bits are 1, and negative otherwise. 
In the complete source distribution, predicting positive when these three bits contain at least two 1s and negative otherwise matches the label on 71.48\% of inputs. 
In OOD, this rule is wrong on every input, reversing the feature--label association while preserving the task rule.

\subsubsection{Model Configuration}
We encode each example with one classification token followed by nine feature tokens. Each token has 64 dimensions: the first 32 form Type, and the remaining 32 form Data. In Type, the first dimension identifies the classification token. For each feature token, one of the next nine dimensions identifies the feature, and one of the following two dimensions encodes its observed binary value. The remaining 20 Type dimensions are zero. The classification token activates only its own indicator.
In Data, the observed binary value of each feature token activates one of the first two dimensions. The remaining 30 dimensions are initialized to zero to zero and serve as scratch space. The classification token's Data is initialized entirely to zero.

The Transformer packs the same observable fields into a width-64 stream. Its single four-head layer uses an FFN with two width-68 hidden layers. Both models predict from the classification token.

\subsubsection{Training Configuration}
Both models use the same 10 examples directly, without augmentation. We train for 2,000 epochs. For paired seeds 41--50, we evaluate the final checkpoint on the remaining 502 examples and report accuracy.

\subsection{SPECT Heart}
The UCI SPECT Heart dataset contains 267 patients represented by 22 binary features extracted from cardiac SPECT images \citep{spect_heart_95}. Its official split contains 80 training and 187 test examples. Our task uses the first nine features and predicts whether at least four are active. 

\subsubsection{Dataset Construction}
\textbf{Training set.}
We select 10 rows from the official 80-row training split, without augmentation.

\textbf{Test set.}
We use the official 187-row test split as the source of the ID test set. We remove the 24 rows whose projected nine-bit inputs also occur in the training set, leaving 163 input-disjoint examples over 91 distinct inputs. Repeated source rows are retained, preserving their original frequencies.

\textbf{OOD test set.}
We construct the OOD test set by shifting the input support from the 112 nine-bit configurations observed in the complete source dataset to configurations not observed in either official split. Specifically, we apply all nine cyclic rotations to each observed configuration and deduplicate the resulting union into 371 inputs. After removing the 112 configurations present in the source dataset, 259 previously unobserved inputs remain, each of which appears once in the OOD test set. Of these, 188 are high-density and 71 are low-density. Cyclic rotations preserve the number of active features and therefore preserve the four-of-nine labeling rule. The test set is OOD because every retained input lies outside the support observed across the complete source dataset, while the labeling rule remains unchanged.

\subsubsection{Model Configuration}
We encode each example with one classification token followed by nine feature tokens. Each token has 64 dimensions: the first 32 form Type, and the remaining 32 form Data. In Type, the first dimension identifies the classification token. For each feature token, one of the next nine dimensions identifies the feature, and one of the following two dimensions encodes its observed binary value. The remaining 20 Type dimensions are zero. The classification token activates only its own indicator.
In Data, the observed binary value of each feature token activates one of the first two dimensions, while the remaining 30 dimensions are initialized to zero and serve as scratch space. The classification token's Data is initialized entirely to zero.

STRAT applies one four-head attention layer and a two-hidden-layer GLU with hidden widths 68 and 68.  The Transformer concatenates the same observable fields into a width-64 residual stream. Its single four-head layer uses a two-hidden-layer FFN with hidden widths 68 and 68. Both models predict from the classification token.

\subsubsection{Training Configuration}
Both models train directly on the same 10 examples, without augmentation. We use full-batch training for 2,000 epochs.

\vspace{-0.1in}
\section{Comparison with State-of-the-Art Methods on Datasets Retaining Their Original Prediction Targets}
\vspace{-0.1in}
\label{app:sota_original_targets}

We compare the STRAT results in Table~\ref{tab:task_overview} with published state-of-the-art results on five datasets retaining their original prediction targets; Appx.~\ref{app:tasks_and_datasets} gives the experimental details. Our augmented training data consist entirely of rule-preserving variants of the same \textbf{10 base examples per dataset}, providing variation within this small source set without introducing additional source instances.

\vspace{-0.1in}
\subsection{SATNet Sudoku}
\vspace{-0.1in}
The Recurrent Transformer of \citet{yang2023recurrent} achieves 100\% exact-board accuracy on $9\times9$ SATNet Sudoku \citep{wang2019satnet}, training on 9,000 puzzles and testing on 1,000. It applies a single four-head Transformer layer recurrently, with 32 steps during training and 64 at evaluation.

STRAT achieves $86.58\pm1.08\%$ exact-board accuracy (Table~\ref{tab:task_overview}). Although this is 13.42 percentage points below the reported perfect score, STRAT trains on only 10 base puzzle--solution pairs, about 0.1\% as many base puzzles, with rule-preserving augmentation (Appx.~\ref{app:tasks_and_datasets}). Our evaluation uses 1,090 held-out puzzles: the official 1,000 test puzzles plus 90 unused puzzles from the training split. Averaged over ten training seeds, STRAT outperforms our Transformer baseline ($27.92\pm4.97\%$), which receives the same training data and augmentation. STRAT also maintains $83.80\pm1.30\%$ accuracy on OOD puzzles with more unevenly distributed clues.

\vspace{-0.1in}
\subsection{Tic-Tac-Toe}
\vspace{-0.1in}
Transformer-based results on Tic-Tac-Toe Endgame are rarely reported as standalone accuracy. TabPFN \citep{hollmann2023tabpfn}, a 12-layer in-context-learning Transformer, reaches 0.9759 ROC-AUC over five 50/50 splits of the 958 boards. CAAFE \citep{hollmann2023caafe} augments TabPFN with GPT-4-generated features and reports validation accuracy rising from 70.0\% to 100\% in a single illustrative run on a 95-instance subsample. Later work \citep{denbreejen2024finetuned,lee2026distpfn} reports only aggregate scores across benchmark suites. The best overall result, 1.000 ROC-AUC, comes from OpenFE with XGBoost \citep{zhang2023openfe,burghardt2026famose}.

STRAT achieves $98.19\pm0.00\%$ mean accuracy (Table~\ref{tab:task_overview}) from only 10 randomly selected base boards, expanded via label-preserving rotations and reflections to 76 training inputs (Appx.~\ref{app:tasks_and_datasets}). This is fewer than one-sixth of the 479 boards TabPFN conditions on. Averaged over ten independent seeds, STRAT outperforms our Transformer baseline($80.91\pm2.53\%$). On an OOD set whose spatial statistics oppose the training correlations, STRAT reaches $94.64\pm0.00\%$, while the Transformer falls to $60.49\pm5.47\%$.
\vspace{-0.1in}
\subsection{Game of Life}
\vspace{-0.1in}
Datasets similar to ours, which pair random toroidal Life boards with their one-step successors under the B3/S23 rule, have been used to evaluate Transformer architectures. On such data, LifeGPT \citep{berkovich2025lifegpt}, a decoder-only GPT with 12 layers and 8 attention heads, trains on 10,000 random $32\times32$ boards and predicts the next state of 109 of 110 held-out broad-entropy boards perfectly. AutomataGPT \citep{berkovich2026automatagpt} extends this setting to about one million trajectories spanning 100 cellular-automaton rules and reaches 98.5\% one-step accuracy on unseen rules. On similar one-step Life data, minimal CNNs rarely converge \citep{springer2020hard}, while a convolutional network with polynomial Kolmogorov--Arnold activations, trained on random $32\times32$ toroidal boards, succeeds in 128 of 128 training runs \citep{ahmed2026polykan}.

STRAT achieves $88.71\pm2.72\%$ exact-board accuracy (Table~\ref{tab:task_overview}) on 757 held-out one-step transitions of $6\times6$ toroidal boards. It trains on only 10 base pairs, 0.1\% of LifeGPT's training set, augmented with random toroidal translations, rotations, and reflections (Appx.~\ref{app:tasks_and_datasets}). Averaged over ten independent seeds, STRAT outperforms our Transformer baseline ($80.54\pm2.64\%$). On 500 OOD boards with fewer live--live adjacencies than any training board, STRAT reaches $79.26\pm12.99\%$, while the Transformer falls to $69.08\pm12.89\%$.
\vspace{-0.1in}
\subsection{PMLB M-of-N}
\vspace{-0.1in}
The strongest reported result on PMLB M-of-N (3,7,10) is 100\% test accuracy, achieved by the Chebyshev Adaptive Network \citep{islam2026bioinspired}. This MLP variant replaces connection weights with degree-$k$ Chebyshev polynomials. For this dataset, \citet{islam2026bioinspired} utilize a degree-3 expansion on a network with a 4-2-output layer structure, reporting the best test accuracy across ten runs. Their dataset comprises 794 examples (roughly 60\% of the 1,324-example PMLB release).

STRAT matches this $100.00\pm0.00\%$ perfect accuracy (Table~\ref{tab:task_overview}) while using fewer than 9\% as many examples as \citet{islam2026bioinspired}. From only 10 randomly selected base examples, expanded via label-preserving cyclic shifts to 70 training inputs (Appx.~\ref{app:tasks_and_datasets}), STRAT achieves zero variance across ten independently trained models (seeds 41--50). Furthermore, STRAT demonstrates perfect out-of-distribution (OOD) generalization ($100.00\pm0.00\%$) on a test set that concentrates both classes at the decision boundary, with negatives having exactly two and positives exactly three active relevant features, an evaluation absent from \citet{islam2026bioinspired}.
\vspace{-0.1in}
\subsection{PMLB MONK-2}
\vspace{-0.1in}
On PMLB MONK-2, \citet{islam2026bioinspired} report 100\% test accuracy using the identical Chebyshev Adaptive Network architecture and best-of-ten evaluation protocol described above. Their dataset includes 360 examples (about 60\% of the 601-example PMLB release). 

STRAT achieves a comparable $90.86\pm3.04\%$ mean accuracy (Table~\ref{tab:task_overview}). Our model is trained from only 10 randomly selected base examples, expanded to 212 unique inputs via label-preserving permutations (Appx.~\ref{app:tasks_and_datasets}). Averaged over ten independent seeds, STRAT successfully maintains its performance under distribution shift, reaching $90.92\pm3.07\%$ when the share of unseen category-count compositions increases from 18.48\% to 32.70\%.

\vspace{-0.1in}
\section{Additional Arithmetic Baselines}
\vspace{-0.1in}
\label{app:arithmetic_baselines}

\paragraph{Sequence adaptations.}
We implement the Neural Accumulator (NAC) and Neural Arithmetic Logic Unit (NALU) modules of \citet{Trask2018} with a single-layer, 64-unit LSTM controller. At each token, affine heads predict the arithmetic matrices from the controller state. The controller receives a one-hot category (literal, plus, or minus) and the numerical value divided by 100; operator values are zero. The arithmetic cell receives the unscaled value and a two-dimensional numerical memory initialized to zero. Its first memory coordinate after the last token predicts the answer. Both models process every real token and freeze their states on padding, with supervision only on the final answer. Parameter counts are 18,700 for NAC and 19,090 for NALU.

\vspace{-0.1in}
\paragraph{Training and reporting.}
The task and training budget follow Section~\ref{sec:locking}. We use full-batch Adam with MSE loss and clip gradient norms at 1. Learning rates are 0.03 for NAC and 0.01 for NALU, selected using development training loss. Both use float64 arithmetic and the final checkpoint. Each run is evaluated on 1,000 OOD expressions. NAC and NALU share per-run training and test samples. Table~\ref{tab:arithmetic_baselines} includes all 400 runs per model. Median training MAEs are 0.34 and 4.02, respectively.

\begin{table}[htbp]
\centering
\caption{Comparison with specialized arithmetic baselines. The Transformer and four STRAT-Lite rows reproduce Table~\ref{tab:locking}; NAC and NALU are evaluated over 400 runs each. P10/P25 denote the 10th/25th percentiles of per-run OOD MAE; the last two columns give the percentage of runs below each error threshold.}
\label{tab:arithmetic_baselines}
\small
\setlength{\tabcolsep}{6pt}
\begin{tabular}{lrrrcc}
\toprule
\midrule
& \multicolumn{3}{c}{OOD MAE $\downarrow$} & \multicolumn{2}{c}{Runs below threshold $\uparrow$} \\
\cmidrule(lr){2-4} \cmidrule(lr){5-6}
Model & P10 & P25 & Median & MAE $<20$ & MAE $<100$ \\
\midrule
Baseline (Config A) & 5,072 & 6,268 & 9,046 & 0\% & 0\% \\
LSTM-controlled NAC & 4,120 & 5,083 & 5,819 & 0\% & 0.25\% \\
LSTM-controlled NALU & 4,973 & 5,327 & 5,770 & 0\% & 0\% \\
\midrule
STRAT Flat (2D/1H) & 33 & 80 & 293 & 5\% & 31\% \\
STRAT Flat MH (2D/2H) & 38 & 97 & 314 & 2\% & 27\% \\
STRAT Compact (3D/1H) & 34 & 77 & 266 & 3\% & 31\% \\
STRAT Std (4D/1H) & \textbf{33} & \textbf{77} & \textbf{258} & 4\% & \textbf{31\%} \\
\midrule
\bottomrule
\end{tabular}
\end{table}

\end{document}